\pdfoutput=1
\def\paperTitle{
Copy the Same, Distill the Difference: \\
Initializing Linear Vision Transformers 
}

\def\authorBlock{%
{\bfseries\fontsize{11.2}{16}\selectfont
    Huaiyuan Qin\,\textnormal{\textsuperscript{1}}%
\ \
    Muli Yang\,\textnormal{\textsuperscript{1}}%
\ \
    Gabriel James Goenawan\,\textnormal{\textsuperscript{1}}%
\ \
    Shiqi Huang\,\textnormal{\textsuperscript{2}}%
}
\vspace{3pt}\\

{\bfseries\fontsize{11.2}{16}\selectfont
    Min Kass Chong\,\textnormal{\textsuperscript{3}}%
\ \
    Wahyu Wiratama\,\textnormal{\textsuperscript{3}}%
\ \
    Peng Hu\,\textnormal{\textsuperscript{4}}%
\ \
    Chen Gong\,\textnormal{\textsuperscript{5}}%
}
\vspace{3pt}\\

{\bfseries\fontsize{11.2}{16}\selectfont
    Wu Liu\,\textnormal{\textsuperscript{6}}%
\ \
    Xi Peng\,\textnormal{\textsuperscript{4}}%
\ \
    Chun Jian Ho\,\textnormal{\textsuperscript{3}}%
\ \
    Hongyuan Zhu\,\textnormal{\textsuperscript{1\,\Letter}}%
}
\vspace{4pt}\\ 

{\fontsize{10.9}{16}\selectfont
\textsuperscript{1}Institute of Advanced Intelligence and Computing (IAIC), A*STAR, Singapore}
\\

{\fontsize{10.9}{16}\selectfont\textsuperscript{2}Nanyang Technological University}
\hspace{3pt}
{\fontsize{10.9}{16}\selectfont\textsuperscript{3}ST Engineering Geo-Insights, Singapore}
\\

{\fontsize{10.9}{16}\selectfont\textsuperscript{4}Sichuan University}
\hspace{3pt}
{\fontsize{10.9}{16}\selectfont\textsuperscript{5}Shanghai Jiao Tong University}
\\

{\fontsize{10.9}{16}\selectfont\textsuperscript{6}University of Science and Technology of China}
\vspace{3pt}\\

{\tt
\small
\texttt{\{qinhy, zhuh\}@a-star.edu.sg}
\hspace{-6.28pt}}\\
}

\newif\ifreview 
\newif\ifarxiv \newcommand{\arxiv}{\arxivtrue}
\newif\ifcamera 

\arxiv

\PassOptionsToPackage{table,dvipsnames}{xcolor}

\documentclass{article}
 
\ifreview \usepackage{meta/iclr_2027} \fi
\ifarxiv \usepackage{meta/iclr_2027} \iclrfinalcopy \fi
\ifcamera \usepackage{meta/iclr_2027} \iclrfinalcopy \fi

\usepackage{fix-cm}
\usepackage{array}
\usepackage{bbm}
\usepackage{nicematrix}

\usepackage{tipa}
\usepackage{dsfont}
\usepackage{etoolbox}  

\usepackage[noend]{algorithmic}
\usepackage{algorithm}

\usepackage{float}
\usepackage{newfloat}
\usepackage{listings}
\floatstyle{ruled}
\newfloat{listing}{tb}{lst}{}
\floatname{listing}{\small Algorithm}

\definecolor{mygray}{RGB}{234,234,234}

\newcommand{\ra}[1]{\renewcommand{\arraystretch}{#1}}

\usepackage{adjustbox}
\usepackage[table,dvipsnames]{xcolor}

\definecolor{darkgreen}{rgb}{0.13, 0.55, 0.13}

\usepackage{graphicx}	
\usepackage{amsmath}	
\usepackage{amssymb}	
\usepackage{booktabs}
\usepackage{times}
\usepackage{epsfig}
\usepackage{caption}
\usepackage{float}
\usepackage{placeins}
\usepackage{color, colortbl}
\usepackage{stfloats}
\usepackage{enumitem}
\usepackage{tabularx}
\usepackage{xstring}
\usepackage{multirow}
\usepackage{xspace}
\usepackage{url}
\usepackage{subcaption}
\usepackage{xcolor}

\usepackage{inconsolata}

\ifcamera \usepackage[accsupp]{axessibility} \fi

% ...

\ifarxiv  \fi

\newcommand{\R}[1]{{%
    \textbf{%
        \ifstrequal{#1}{1}{\textcolor{red}{R#1}}{%
        \ifstrequal{#1}{2}{\textcolor{blue}{R#1}}{%
        \ifstrequal{#1}{3}{\textcolor{magenta}{R#1}}{%
        \ifstrequal{#1}{4}{\textcolor{teal}{R#1}}{%
                           \textcolor{cyan}{R#1}%
        }}}}%
    }%
}}

\definecolor{Gray}{gray}{0.5}
\definecolor{nicergreen}{rgb}{0.13, 0.54, 0.13}
\definecolor{nicered}{rgb}{0.83, 0.16, 0.16}
\definecolor{lightgray}{RGB}{230, 230, 230}
\definecolor{Highlight}{HTML}{39b54a}  %

\usepackage{appendix}
\usepackage{footnote}

\usepackage{microtype}
\usepackage{cuted}
\usepackage{tocloft}

\usepackage[T1]{fontenc}
\usepackage{DejaVuSans}

\usepackage{ifsym, marvosym}

\let\svthefootnote\thefootnote
\newcommand\freefootnote[1]{%
  \let\thefootnote\relax%
  \footnotetext{#1}%
  \let\thefootnote\svthefootnote%
}

\makeatletter
\DeclareRobustCommand\onedot{\futurelet\@let@token\@onedot}
\def\@onedot{\ifx\@let@token.\else.\null\fi\xspace}

 \def\vs{\emph{vs}\onedot}

\makeatother

\usepackage{wrapfig}  

\usepackage{xr-hyper}

\makeatletter
\newcommand*{\addFileDependency}[1]{
  \typeout{(#1)}
  \@addtofilelist{#1}
  \IfFileExists{#1}{}{\typeout{No file #1.}}
}

\makeatother

\definecolor{cvprblue}{rgb}{0.21,0.49,0.74}
\usepackage[pagebackref,breaklinks,colorlinks]{hyperref}
\usepackage[capitalize]{cleveref}
\crefname{section}{Sec.}{Secs.}
\crefname{table}{Table}{Tables}
\crefname{figure}{Fig.}{Figs.}

\usepackage[utf8]{inputenc} 
\usepackage[T1]{fontenc}    
\usepackage{url}            
\usepackage{booktabs}       
\usepackage{amsfonts}       
\usepackage{amssymb}        
\usepackage{nicefrac}       
\usepackage{microtype}      
\usepackage{xcolor}         

\title{\paperTitle}
\author{\authorBlock}

\begin{document}
\maketitle

\ifarxiv
\lhead{Preprint.}
\freefootnote{
\hspace{-12pt}
\textsuperscript{\Letter}Corresponding author.
}
\fi

\begin{abstract}
Linear Vision Transformers (ViTs) are designed to replace the attention in Softmax ViTs with the linear-complexity attention operator for more efficient token routing, but they require from-scratch pre-training and typically underperform the original Softmax version.
How to initialize linear ViTs both efficiently and effectively still remains unclear.
In this work, we explicitly ask: given that most foundation ViTs are built on the mainstream Softmax attention, can linear ViTs benefit from their pre-trained weights?
Recent works on Attention Transfer show that attention is the effective transferable component between Softmax ViTs, suggesting attention alone suffices for such reuse.
However, we find the opposite for Softmax-to-linear transfer.
The attention weights are operator-specific: copying them barely helps, and is sometimes even worse than random initialization.
Instead, the attention's token routing behavior can be recovered through distillation with a proper loss design, 
letting linear ViTs reduce the gap and even match Softmax ones.
In contrast, the MLP weights, which carry the learned representation, are operator-agnostic: they can be transferred by simple direct copying, which already carries most of the benefit of the pre-trained weights.
Thus, copying MLPs can serve as an effective foundation for Softmax-to-linear transfer: paired with the distilled attention, linear ViTs eventually close the remaining gap and even surpass Softmax ones.
These findings hold consistently across various linear ViT variants, different model sizes, and diverse datasets.
We hope this study deepens the understanding of reusing pre-trained weights across attention operators: copy what stays the same and distill what differs, to recover the benefit across the Softmax-to-linear boundary.
\end{abstract}
\section{Introduction}
\label{sec:intro}

Vision Transformers (ViTs)~\citep{dosovitskiy2020image} have become the dominant backbone for visual representation learning in the modern era, delivering strong performance across a wide range of vision tasks.
Their Softmax attention mechanism, however, incurs a quadratic $\mathcal{O}(N^2)$ cost in the number of tokens, making token routing a computational bottleneck with high-resolution or long-context tasks.
To reduce this cost, linear ViTs are designed to replace the attention in Softmax ViTs with the linear-complexity attention operator for more efficient token routing, but they require from-scratch pre-training and typically underperform the original Softmax version.
How to initialize linear ViTs efficiently and effectively is therefore a question of practical significance, yet still remains unclear.

\begin{figure}[t]
    \centering
    \includegraphics[width=\linewidth]{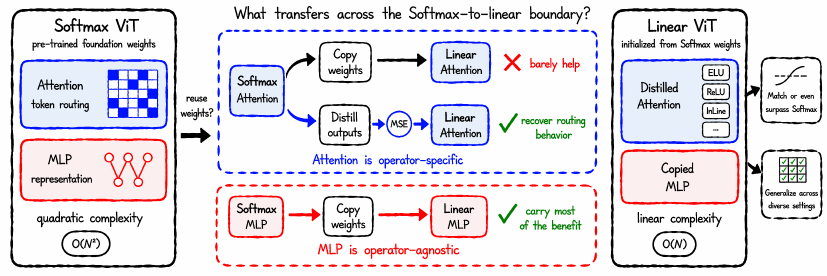}
    \vspace{-15pt}
    \caption{\textbf{What survives the Softmax-to-linear boundary?}
    We ask whether linear ViTs can benefit from pre-trained Softmax weights for initialization.
    We find that the attention weights are \emph{operator-specific} that copying them barely helps.
    Their token routing behavior can instead be recovered by distilling the attention outputs, while the \emph{operator-agnostic} MLP weights, which carry the learned representation, can be simply transferred by direct copying.
    Paired together, linear ViTs can match and even surpass the Softmax ones, consistently across diverse settings.}
    \vspace{-15pt}
    \label{fig:teaser}
\end{figure}

Therefore, in this work, we explicitly ask: given that most foundation ViTs are built on the mainstream Softmax attention, can linear ViTs benefit from their pre-trained weights for initialization?
Analytical initializations~\citep{trockman2023mimetic,zheng2025structured} have shown the potential to improve from-scratch pre-training, but they still remain well below vanilla Softmax ViTs.
More directly, recent works on Attention Transfer~\citep{li2024attention,qin2026attention} have demonstrated that, \emph{between Softmax ViTs}, transferring
only the attention, by copying or distilling its attention maps, can already recover the original model's performance, suggesting Softmax attention alone suffices for such reuse.

However, we find the \textit{\textbf{opposite}} for Softmax-to-linear transfer.
We test the most direct reuse strategy for initialization: copying the pre-trained Softmax attention projection weights into a linear ViT, while leaving all other components randomly initialized.
As shown in~\Cref{fig:reversal_grid}, such simple copy-only transfer from Softmax to linear ViTs barely helps, and is sometimes even worse than pure random initialization.
Such failure is consistent across the linear ViT variants we study, revealing that the pre-trained attention weights are \emph{operator-specific}: copied on their own, their benefit does not survive the change of the attention operator.

Does this failure mean the Softmax attention is entirely useless for linear ViTs?
Crucially, no.
While its weights do not transfer by copying, we identify that the token routing behavior of Softmax attention can be recovered through distillation with a proper loss design: unlike the map matching in~\citet{li2024attention}, the loss should match the attention outputs rather than the operator-specific attention maps.
This output-matching distillation lets linear ViTs largely reduce the gap to Softmax ones and even match them.
The practical message is therefore not to copy the attention, but to distill it instead.

Furthermore, if the attention should be distilled rather than copied, what remains worth copying?
To answer this, we decompose the transfer at the component level, selectively initializing the attention and the MLP blocks of linear ViTs from the pre-trained Softmax weights.
We uncover that the MLP blocks are \emph{operator-agnostic}: simply copying them already carries most of the benefit of the pre-trained weights. 
What transfers effectively across the Softmax-to-linear boundary by copying is the representation in the MLP blocks, consistent with the findings in~\citet{geva2021transformer}.
Moreover, we find that copying MLPs can serve as an effective foundation for Softmax-to-linear transfer: when paired with distilled attention, linear ViTs eventually close the remaining gap and can even surpass Softmax ones; whereas copied or randomly initialized attention still leaves them below Softmax level.

We systematically validate these findings across various linear ViT variants, different model sizes, and diverse datasets.
Across these settings, a simple principle emerges (see~\Cref{fig:teaser}): copy what stays the same and distill what differs, to recover the benefit across the Softmax-to-linear boundary.
We hope this principle can deepen the understanding of reusing pre-trained weights across attention operators.

In summary, our primary contributions are as follows:
\begin{enumerate}
    \item \textbf{Systematic Study of Softmax-to-Linear Transfer.} We present the first systematic component-level study of initializing linear ViTs from pre-trained Softmax weights across linear attention operators, and reveal the attention weights are \textit{operator-specific}: copying them barely helps and can fall below random initialization.
    \item \textbf{Distillation as the Fix.} We show that a properly designed distillation loss, matching attention outputs rather than attention maps, recovers the token routing behavior and lets linear ViTs close most of the gap or even match the corresponding Softmax ones.
    \item \textbf{A Simple, General Recipe.} We find that copying the \textit{operator-agnostic} MLP weights carries most of the transfer benefit, and combining them with distilled attention can let linear ViTs surpass Softmax, yielding the principle: copy what stays the same and distill what differs.
\end{enumerate}


\section{Related Work}
\label{sec:related_work}

\paragraph{Efficient Linear ViTs.}
Linear ViTs replace the attention in Softmax ViTs with the linear-complexity attention operator for more efficient token routing, stemming from the kernelized linear attention introduced in~\citet{katharopoulos2020transformers}.
Recent vision designs further diversify this family:
InLine~\citep{han2024bridging} restores injectivity through subtraction normalization,
MHLA~\citep{zhang2026mhla} restores expressivity through token-level multi-head computation over spatial blocks,
and TTT~\citep{han2026vit3} replaces explicit attention with learned test-time states.
This direction also extends broadly across modalities: focused linear attention and circulant attention for vision~\citep{han2023flatten,han2026circulant}, gated linear attention and delta-rule parallelization for language modeling~\citep{yang2024gated,yang2024parallelizing}, and state-space models (SSMs) as an alternative linear-complexity route~\citep{gu2024mamba,liu2024vmamba}.
Recent analysis further shows that a broad class of TTT architectures can be reduced to an implicit linear-attention operator~\citep{liu2026ttt}, supporting a unified view of these models.
However, these designs typically rely on dedicated pre-training from scratch, and whether they can instead be initialized from pre-trained Softmax ViTs has not been systematically studied.

\paragraph{Initialization for ViTs.}
Beyond default random initialization, analytical methods initialize weights from structural priors: 
Mimetic~\citep{trockman2023mimetic} shapes the query-key product toward identity-like attention, with extensions to MLPs~\citep{trockman2026mimetic} and SSMs~\citep{trockman2024ssm}, 
while structured initialization~\citep{zheng2025structured} imposes convolution-like attention patterns.
Another line of work aims to initialize smaller models directly from the weights of larger pre-trained ones~\citep{xu2024initializing}.
These methods, however, either construct weights without any pre-trained source or reuse weights within the same attention operator;
we instead study reusing pre-trained weights across the attention-operator boundary.

\paragraph{Attention Transfer for ViTs.}
Within the broader framework of knowledge distillation~\citep{hinton2015distilling}, the idea of Attention Transfer has been extended to ViTs~\citep{wang2022attention}.
Recently,~\citet{li2024attention} showed that attention patterns alone can suffice to recover the benefit of pre-trained weights between Softmax ViTs, and~\citet{qin2026attention} further identified the validity boundary of such transfer under teacher-student architectural mismatch, still within the Softmax family.
However, whether such attention-only transfer remains effective across the attention-operator boundary has not been studied.
Our study answers this question by splitting such transfer at the component level: attention weights are operator-specific, whereas MLP weights are operator-agnostic.

\paragraph{Linearizing Transformers.}
A parallel line of work studies how to convert pre-trained Transformers into linearized ones.
In language modeling, Hedgehog~\citep{zhang2024hedgehog} learns kernels to mimic the original attention maps, whereas LoLCATs~\citep{zhang2025lolcats} and RADLADS~\citep{goldstein2025radlads} instead match the attention outputs when scaling such conversion to large pre-trained models.
Related efforts have also appeared in vision: 
ViT-AdaLA~\citep{li2026vit} adapts pre-trained Softmax weights to linear ViTs,
\citet{wei2025vit, wang2025data} distill pre-trained ViTs into SSMs, 
T$^5$~\citep{li2026linearizing} converts a pre-trained ViT into the TTT architecture,
DiD~\citep{qin2026did} extends label-free conversion to object detection by aligning detector-facing interfaces,
and similar conversions have also been applied to linear diffusion transformers for more efficient image generation~\citep{liu2024clear,wang2025lit}.
These methods show that Softmax-to-linear transfer is feasible, but each relies on dedicated adaptation before downstream use.
Our study instead offers a simple, operator-general recipe: copy the operator-agnostic MLPs and distill the attention outputs.

\vspace{-3pt}

\section{Preliminaries}
\label{sec:preliminary}

\subsection{From Softmax to Linear Attention}

\paragraph{Softmax Attention.}
Given the queries $Q$, keys $K$, and values $V$ at block $\ell$, a Softmax ViT~\citep{dosovitskiy2020image} routes tokens through the attention function\footnote{We omit the scaling factor $1/\sqrt{d_k}$ for simplicity.}:
\begin{align}
\label{eq:softmax_attn}
    f_\text{softmax}^{(\ell)} = \mathrm{softmax}\left(Q^{(\ell)}K^{(\ell)\top}\right)V^{(\ell)},
\end{align}
which computes an \emph{attention map} at a quadratic $\mathcal{O}(N^2)$ cost in the number of tokens $N$.

\paragraph{Linear Attention.}
Linear ViTs replace this Softmax operator with a linear-complexity alternative,
in its standard form via a kernel function $\phi(\cdot)$ that reorders the computation as:
\begin{align}
\label{eq:linear_attn}
    f_\text{linear}^{(\ell)} = \frac{\phi\big(Q^{(\ell)}\big)\left(\phi\big(K^{(\ell)}\big)^\top V^{(\ell)}\right)}{\phi\big(Q^{(\ell)}\big)\left(\phi\big(K^{(\ell)}\big)^\top \mathbf{1}_N\right)}.
\end{align}
The computational cost is therefore reduced to $\mathcal{O}(N)$ by the multiplication reordering. 

\subsection{Experimental Setup}
\label{subsec:exp_setup}

\paragraph{Model Zoo.}
We study five representative linear ViT variants:
ELU and ReLU~\citep{katharopoulos2020transformers}, which instantiate $\phi$ as $\mathrm{ELU}(\cdot)+1$ and $\mathrm{ReLU}(\cdot)$, respectively;
InLine~\citep{han2024bridging}, which uses an identity kernel with subtraction-based normalization;
MHLA~\citep{zhang2026mhla}, which restores the expressivity of linear attention through token-level multi-head computation with auxiliary convolutions;
and TTT~\citep{han2026vit3}, which replaces the attention computation entirely with a compact linear-complexity inner model constructed from KV pairs.
The first three variants are drop-in linear operators whose weight shapes exactly match the Softmax attention, while MHLA and TTT introduce additional components.
Full descriptions of different linear ViTs and the corresponding attention operators are given in~\Cref{supp:model_zoo}.

\paragraph{Transfer Protocol.}
To conduct a systematic analysis of the Softmax-to-linear attention transfer, we instantiate the five linear ViT variants across Tiny, Small, and Base sizes, all with the DeiT~\citep{touvron2021training} pre-trained weights of the corresponding model size as the transfer source.
Unless specifically mentioned, all linear ViTs are with their default implementations and hyper-parameters.
For linear ViTs with additional components beyond the Softmax ones, the unmatched parameters are initialized following their default settings.
More details are in~\Cref{supp:transfer_protocol}.

\paragraph{Evaluation Protocol.}
We evaluate Tiny and Small linear ViTs on six transfer datasets: CIFAR-10, CIFAR-100~\citep{krizhevsky2009learning}, STL-10~\citep{coates2011analysis}, Food~\citep{bossard2014food}, Flowers~\citep{nilsback2008automated}, and Pets~\citep{parkhi2012cats}.
We additionally evaluate all three sizes on ImageNet-1K~\citep{deng2009imagenet} for large-scale validation.
We conduct all the experiments with the default settings following the recipe of~\citet{xu2024initializing}, training for 300 epochs or 600 epochs depending on the dataset scale.
All reported results are averaged over 3 random seeds to ensure statistical significance.
More implementation details are provided in~\Cref{supp:eval_protocol}.

\paragraph{Reference Baselines.}
For reference purposes, we set two baselines across all three model sizes for comparison. 
\emph{Random} baseline defines the from-scratch reference of the Softmax-to-linear transfer: one transfer is considered effective only if it improves over training from scratch.
\emph{Softmax} baseline defines the recovery target: the performance of the pre-trained Softmax ViT of the same model size on the same task, which the transfer aims to recover.


\section{Softmax Attention Weights Are Operator-Specific}
\label{sec:main_results}

\begin{figure*}[t]
    \centering
    \includegraphics[width=\linewidth]{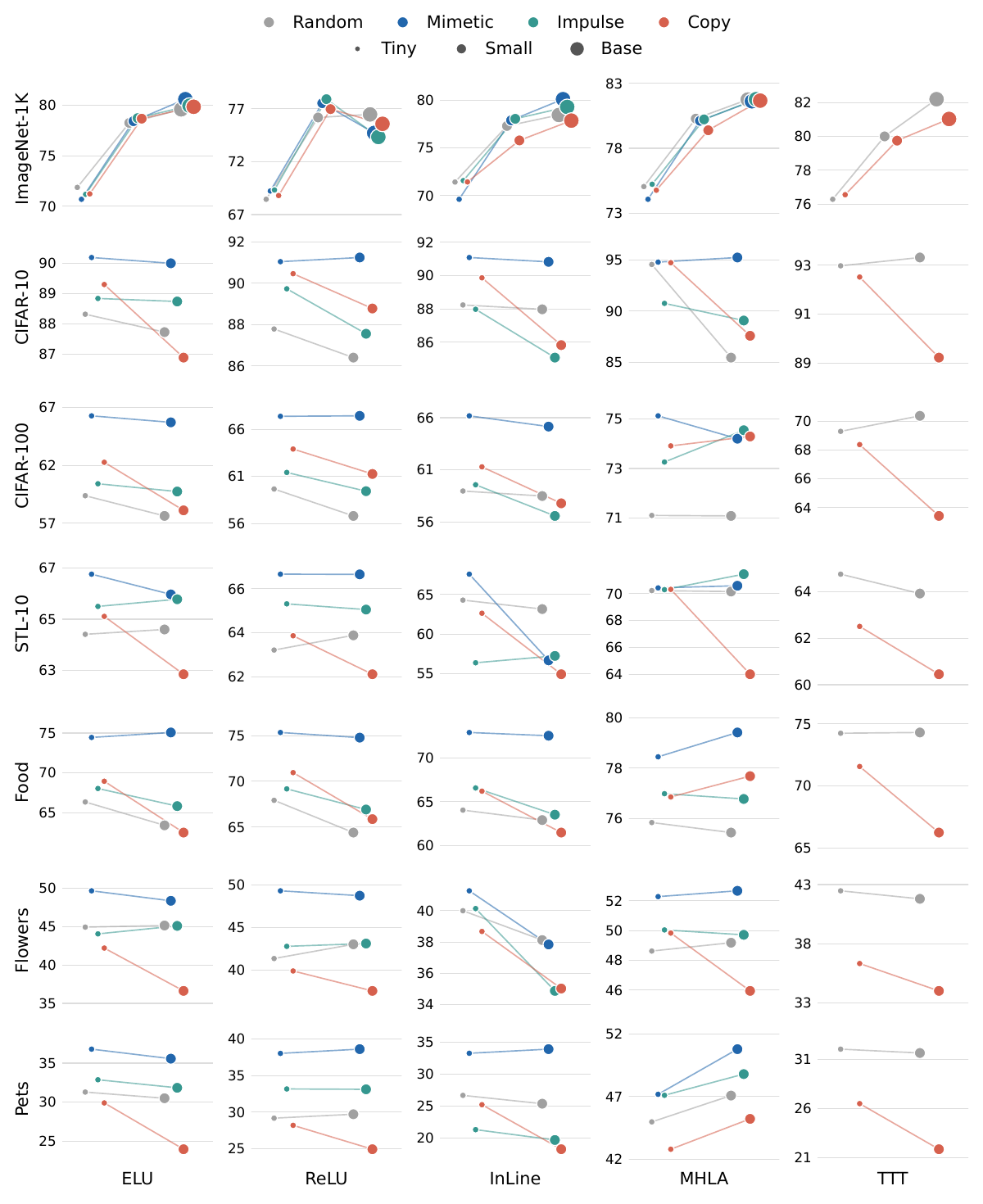}
    \vspace{-18pt}
    \caption{\textbf{Attention-only Softmax-to-linear transfer barely helps.}
    We begin by evaluating the attention-only initializations across five linear ViTs (columns) and seven datasets (rows).
    We report the Top-1 Accuracy (\%); marker size denotes the model size (Tiny/Small/Base), and lines denote the same initialization across sizes.
    Across all operators, datasets, and sizes, \textit{Copy} never stands out as the strongest attention-only initialization and remains at or near the \textit{Random}-initialization tier.
    }
    \vspace{-12pt}
    \label{fig:reversal_grid}
\end{figure*}

We begin by asking whether the token routing behavior encoded in pre-trained Softmax attention weights can be simply transferred as weights by copying across the Softmax-to-linear boundary.

\subsection{Attention-Only Transfer}
\label{subsec:attn_only_setup}
We first evaluate the strategy motivated by Attention Transfer between Softmax ViTs~\citep{li2024attention,qin2026attention}: transferring only the attention.
Beyond the reference baselines defined in~\Cref{subsec:exp_setup}, we further include additional weight-initialization settings as follows.
\textit{Copy} baseline loads the pre-trained Softmax attention projection weights, including queries $Q$, keys $K$, values $V$, into every block of the linear ViT, while leaving all other components randomly initialized.
\textit{Mimetic}~\citep{trockman2023mimetic} and \textit{Impulse}~\citep{zheng2025structured} are two analytical initializations that construct attention weights from structural priors without using any pre-trained source, serving as stronger baselines than \textit{Random}.\footnote{Mimetic and Impulse assume the standard $QK^\top$ attention parameterization and are therefore incompatible with TTT~\citep{han2026vit3}.}
We compare all these baselines under the same training recipe.
The assumption is that if the pre-trained Softmax attention alone suffices to preserve model performance across the Softmax-to-linear boundary, \textit{Copy} should provide a clear advantage over all three alternatives.

\subsection{Main Results}
\label{subsec:reversal_main}

\Cref{fig:reversal_grid} reports the comparison across all five linear ViTs, seven datasets, and different model sizes.
The results reveal a clear failure of the expected advantage of \textit{Copy}.
Across all the experiments, simply copying the Softmax weights into the linear operator, which has been shown effective in~\citet{li2024attention,qin2026attention} between Softmax ViTs, 
\textit{never} stands out as the strongest initialization.

This observed failure appears mainly in the following three ways:
First, adopting Softmax weights loses to pre-training-free initialization: across datasets, \textit{Mimetic}, built without ever using any pre-trained source, systematically outperforms \textit{Copy}, which loads the pre-trained Softmax attention weights.
Second, copying barely helps: even against \textit{Random}, \textit{Copy} yields no consistent gain, in sharp contrast to the recovery attention-only transfer achieves between Softmax ViTs.
Third, copying can sometimes hurt: as shown, \textit{Copy} degrades performance by large margins compared to \textit{Random}; this drop is consistent for TTT and also appears for other linear ViTs on datasets such as Pets and Flowers.
Taken together, these results show that direct attention-only copying provides no reliable initialization advantage across the Softmax-to-linear boundary, the opposite of the transfer behavior observed between Softmax ViTs.
Further interpretations of this failure are given in~\Cref{supp:transfer_explanation}.

\begin{table}[t]
    \begin{minipage}[t]{0.48\textwidth}
       \centering
\small
\caption{\textbf{Pre-trained weights robustness.}
We repeat the attention-only transfer on ViT-Base with different pre-trained Softmax weights, under the same ImageNet-1K evaluation.
\textcolor{gray}{Gray row} shows the shared \textit{Random} baseline, and colored deltas indicate gains/drops relative to it.
The failure remains consistent across all pre-trained sources: none provides a reliable advantage over random initialization.
}
\vspace{-0.8pt}
\label{tab:pretrain_robustness}
\ra{1.02}
\setlength{\tabcolsep}{3.2pt}
\footnotesize
\begin{tabular}{l cccc}
 & ELU & ReLU & MHLA & TTT \\
\midrule
\color{gray} Random & \color{gray} 79.6 & \color{gray} 76.5 & \color{gray} 81.8 &  \color{gray} 82.2 \\
DeiT  & 79.8 {\scriptsize\color[HTML]{2166ac}+0.2} & 75.6 {\scriptsize\color[HTML]{d6604d}-0.9} & 81.7 {\scriptsize\color[HTML]{d6604d}-0.1} & 81.0 {\scriptsize\color[HTML]{d6604d}-1.2} \\
DINO    & 80.2 {\scriptsize\color[HTML]{2166ac}+0.6} & 75.2 {\scriptsize\color[HTML]{d6604d}-1.3} & 81.6 {\scriptsize\color[HTML]{d6604d}-0.2} & 80.6 {\scriptsize\color[HTML]{d6604d}-1.6} \\
MoCov3 & 79.9 {\scriptsize\color[HTML]{2166ac}+0.3} & 75.7 {\scriptsize\color[HTML]{d6604d}-0.8} & 81.7 {\scriptsize\color[HTML]{d6604d}-0.1} & 81.1 {\scriptsize\color[HTML]{d6604d}-1.1} \\
iBOT         & 80.3 {\scriptsize\color[HTML]{2166ac}+0.7} & 75.3 {\scriptsize\color[HTML]{d6604d}-1.2} & 81.8 {\scriptsize\color{gray}+0.0} & 80.7 {\scriptsize\color[HTML]{d6604d}-1.5} \\
MAE          & 80.4 {\scriptsize\color[HTML]{2166ac}+0.8} & 76.0 {\scriptsize\color[HTML]{d6604d}-0.5} & 82.0 {\scriptsize\color[HTML]{2166ac}+0.2} & 81.4 {\scriptsize\color[HTML]{d6604d}-0.8} \\
\end{tabular}

    \end{minipage}
    \hfill
    \begin{minipage}[t]{0.48\textwidth}
       \centering
\small
\caption{\textbf{Out-of-distribution robustness.}
We evaluate the transferred ViT-Base models on four distribution-shift benchmarks.
Colored deltas indicate gains/drops relative to the \textcolor{gray}{\textit{Random}} baseline, and \textit{Softmax} denotes the Softmax target reference.
Copying the Softmax attention gains no additional robustness and stays equally far from the Softmax reference, consistent with the in-distribution evaluation.
}
\vspace{3.4pt}
\label{tab:ood_reversal}
\ra{1.02}
\setlength{\tabcolsep}{1.6pt}
\footnotesize
\begin{tabular}{l cc cc c}
 & \multicolumn{2}{c}{ELU} & \multicolumn{2}{c}{TTT} & \multirow{2}{*}{Softmax}\\
\cmidrule(lr){2-3} \cmidrule(lr){4-5}
 & \color{gray}Random & Copy & \color{gray}Random & Copy & \\
\midrule
IN-A     & \color{gray}18.1 & 19.1 {\scriptsize\color[HTML]{2166ac}+1.0} & \color{gray}24.1 & 23.7 {\scriptsize\color[HTML]{d6604d}-0.4} & 29.2 \\
IN-R      & \color{gray}38.9 & 39.6 {\scriptsize\color[HTML]{2166ac}+0.7} & \color{gray}42.3 & 42.2 {\scriptsize\color[HTML]{d6604d}-0.1} & 45.8 \\
IN-S & \color{gray}27.2 & 27.8 {\scriptsize\color[HTML]{2166ac}+0.6} & \color{gray}30.0 & 29.9 {\scriptsize\color[HTML]{d6604d}-0.1} & 33.4 \\
IN-V2     & \color{gray}66.8 & 67.2 {\scriptsize\color[HTML]{2166ac}+0.4} & \color{gray}69.3 & 69.1 {\scriptsize\color[HTML]{d6604d}-0.2} & 72.4 \\
\end{tabular}

    \end{minipage}
    \vspace{-15pt}
\end{table}

\subsection{Robustness of the Failure}
\label{subsec:reversal_robustness}

One natural counter-hypothesis is that this observed failure might be an artifact of the evaluated setting: with larger model capacity, with Softmax weights from a different pre-training algorithm, or beyond the in-distribution evaluation setting, copying might still be able to deliver its expected advantage.
To rule this out, we rigorously test the robustness of this failure along the three axes below.

\paragraph{Model Size Robustness.}
We first compare the attention-only initializations across the evaluated sizes on all transfer datasets.
As shown in~\Cref{fig:reversal_grid}, the qualitative failure pattern holds consistently at all the tested sizes: within each plot, the attention-only \textit{Copy} remains at or near the \textit{Random}-initialization tier.
In contrast, the analytical initializations hold their advantage as the model size grows, further confirming that the failure is robust to model size and is not an artifact of limited model capacity.

\paragraph{Pre-trained Weights Robustness.}
We then test whether the failure is merely a deficiency of the supervised DeiT weights.
\Cref{tab:pretrain_robustness} reports the attention-only transfer with the Softmax weights from other well-known pre-training algorithms, including DINO~\citep{caron2021emerging}, MoCov3~\citep{chen2021empirical}, iBOT~\citep{zhou2021ibot}, and MAE~\citep{he2022masked}, under the same ImageNet-1K evaluation.
The same trend appears for all weights: copying can shift the performance only marginally, but can never turn into a reliable advantage over \textit{Random}.
This confirms that the failure is robust to the choice of pre-trained weights, while changing this Softmax source cannot reverse it.

\paragraph{Out-of-Distribution Robustness.}
The observed failure also persists under the out-of-distribution evaluation.
\Cref{tab:ood_reversal} further tests the transferred models on four ImageNet-1K distribution-shift benchmarks~\citep{hendrycks2021natural,hendrycks2021many,wang2019learning,recht2019imagenet}, where \textit{Copy} merely preserves its marginal in-distribution difference from \textit{Random} and gains no additional robustness, while both initializations remain equally far from the Softmax reference.
This further confirms that the failure is robust to substantial distribution shift.

Overall, the comprehensive evaluations above show that attention-only Softmax-to-linear copy consistently stays at or near the random-initialization tier.
Thus, the pre-trained Softmax attention weights are \emph{operator-specific}: copied on their own, their benefit does not survive the change of the attention operator.
This raises the questions: does this failure mean that the Softmax attention is entirely useless for linear ViTs, and if its weights should not be copied, what remains worth copying?

\section{Distill the Routing, Copy the Representation}
\label{sec:localize}

Having established the failure of attention-only copying, we now ask what exactly the pre-trained Softmax weights can offer to linear ViTs.
We answer the two questions raised in~\Cref{sec:main_results} in turn: the attention's token routing behavior can be effectively transferred through distillation rather than copying, while the representation carried in the MLP blocks can be transferred through direct copying; together, the two findings form a simple recipe for Softmax-to-linear transfer.
Unless specifically mentioned, all results below are reported with ViT-Small on the transfer datasets.

\begin{figure}[t]
    \centering
    \includegraphics[width=\linewidth]{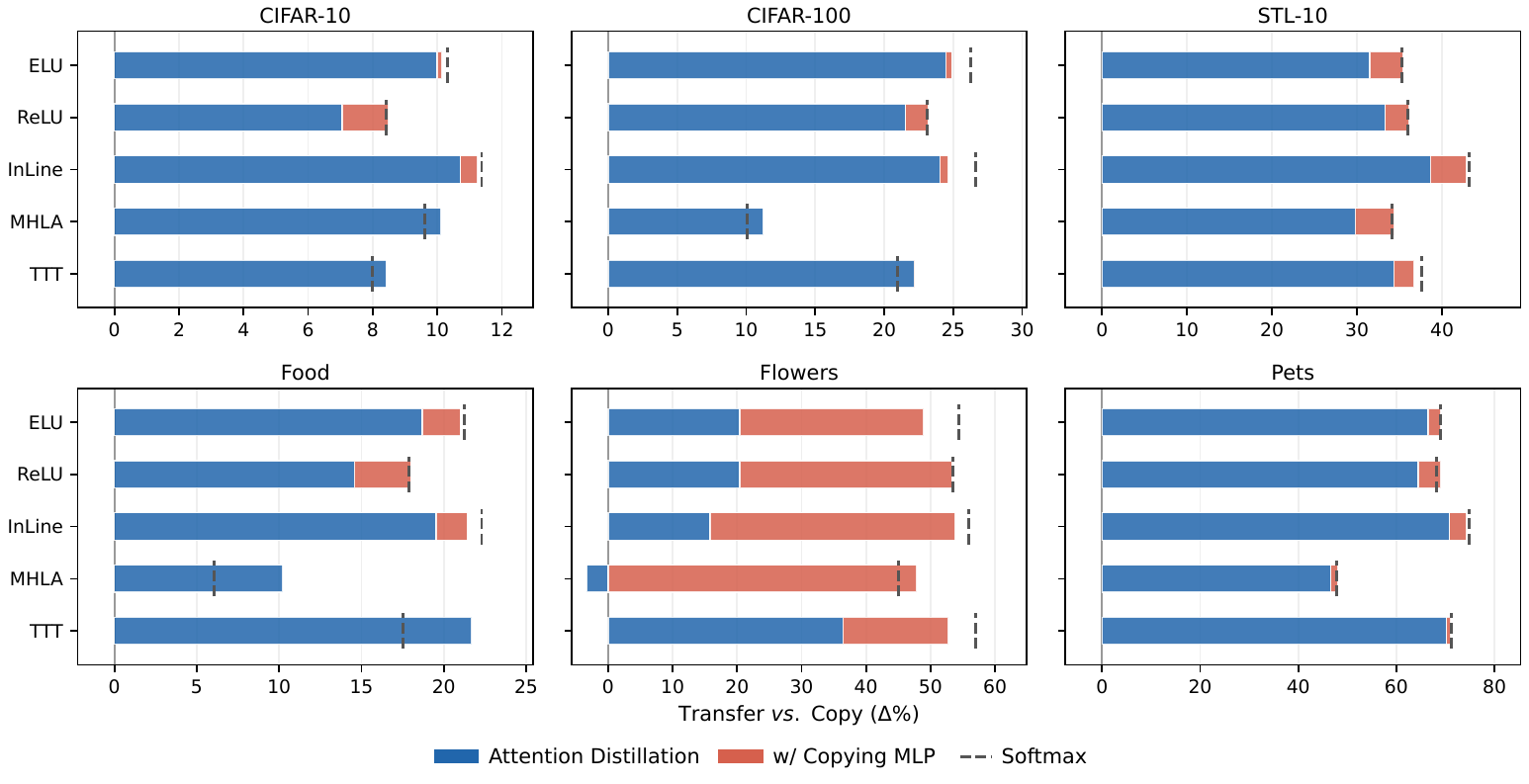}
    \vspace{-18pt}
    \caption{\textbf{Distillation rescues attention-only copying.}
    For each linear operator with ViT-Small, we report the accuracy gain over \textit{Copy}:
    \textcolor[HTML]{2166ac}{blue} bars show the gain from distilling the attention on top of \textit{Copy}, and \textcolor[HTML]{d6604d}{red} bars are the additional gain from further copying the MLP weights (\textit{FullCopy}), leading the full setting to correspond to \textit{FullCopy w/ Distill}.
    The dashed line marks the corresponding Softmax recovery target.
    Distillation effectively recovers most of the gap across linear operators on most datasets; further combining the copied MLPs can even reach the Softmax target.
    }
    \vspace{-12pt}
    \label{fig:distill_rescue}
\end{figure}

\subsection{Token Routing Can Be Recovered through Distillation}
\label{subsec:distill}

Given the failure of direct copying, we hypothesize that the token routing behavior encoded in the Softmax attention can be transferred as \emph{behavior} instead of as weights.
To test this, we distill from a frozen pre-trained Softmax teacher during downstream training: an additional distillation loss aligns the attention behavior between the student and the teacher at every block, with $\lambda$ as the weight of this loss in the overall training objective.
For transfer between Softmax ViTs, this distillation loss is computed between the attention maps as in~\citet{li2024attention,qin2026attention}.
However, for Softmax-to-linear transfer, given the differences among linear operators and the resulting operator-specific attention maps, with some operators never materializing a map at all, we deliberately place the loss between the attention \emph{outputs}.
This output-matching choice parallels the practice in language-model linearization~\citep{zhang2025lolcats}, in contrast to map-based mimicry~\citep{zhang2024hedgehog}.

\Cref{fig:distill_rescue} shows that this distillation effectively rescues the attention-only copying across linear operators on various datasets.\footnote{All reported distillation results in \Cref{fig:distill_rescue} use the output-matching loss with a fixed calibrated $\lambda=50$, shared across all linear operators and datasets. 
A detailed analysis of the loss design is given in~\Cref{subsec:loss_analysis}.
}
Compared to \textit{Copy}, attention distillation closes most of the gap to the Softmax recovery target for every linear operator, and can even reach this target for MHLA and TTT on some datasets.
One exception is the fine-grained Flowers, where attention distillation alone leaves a larger gap, and for MHLA it even falls below \textit{Copy}:
with the small number of training samples, the task signal appears too weak to relearn the representation that attention-only supervision cannot supply.

Together with the failure of copying established in~\Cref{sec:main_results}, these results confirm that the token routing does transfer across the Softmax-to-linear boundary: as behavior through distillation, not as weights through copying.
The practical message for the attention is thus not to copy it, but to distill it instead.
This directly motivates the second question: if the attention should be distilled rather than copied, what of the pre-trained weights remains worth copying?

\begin{figure}[t]
    \centering
    \includegraphics[width=\linewidth]{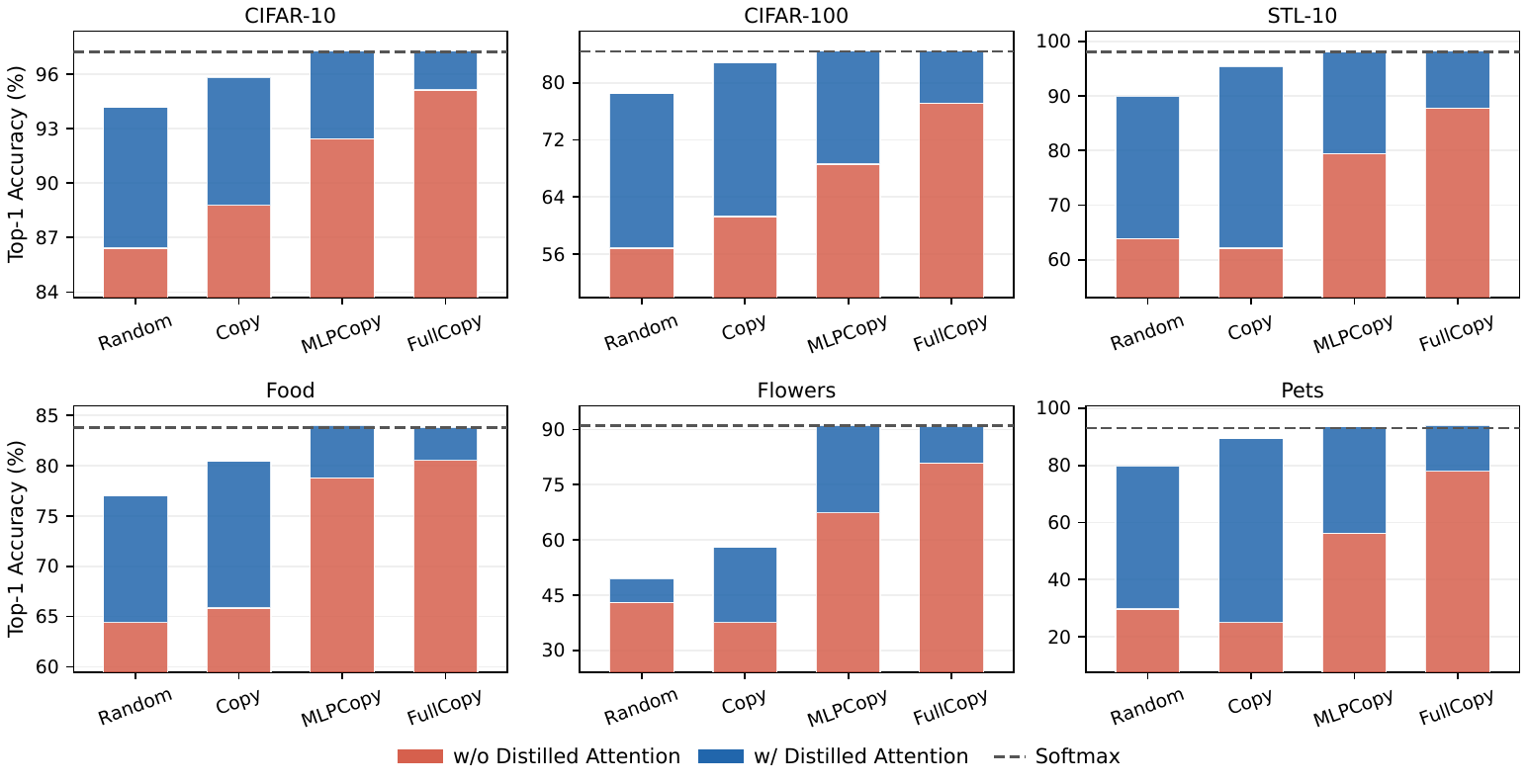}
    \vspace{-18pt}
    \caption{\textbf{MLP weights survive Softmax-to-linear transfer.}
    We selectively initialize the ReLU-based linear ViT-Small by copying from the pre-trained Softmax weights at the component level, yielding four weight-initialization settings: \textit{Random}, \textit{Copy}, \textit{MLPCopy}, and \textit{FullCopy}.
    For each, we report its plain accuracy and the accuracy with the distilled attention (\textit{w/ Distill}).
    \textcolor[HTML]{d6604d}{Red} bars denote the accuracy of each initialization alone, and \textcolor[HTML]{2166ac}{blue} bars show the additional gain from the distilled attention.
    The dashed line marks the corresponding Softmax recovery target.
    Copying MLPs can already effectively carry most of the benefit of the pre-trained weights, while distilling the attention can lift every one of them and bring \textit{MLPCopy} and \textit{FullCopy} to match the Softmax level.
    }
    \vspace{-12pt}
    \label{fig:factorial}
\end{figure}

\subsection{MLP Weights Are Operator-Agnostic}
\label{subsec:ffncopy}

To answer this, we decompose the transfer at the component level by selectively initializing the attention and the MLP blocks of linear ViTs.
We additionally define two more weight-initialization settings:
\textit{MLPCopy}, which directly copies the MLP blocks from the pre-trained Softmax weights, while leaving all attention-related components randomly initialized;  
and \textit{FullCopy}, which copies all pre-trained Softmax weights.
\Cref{fig:factorial} shows a direct comparison of the four weight-initialization settings with the ReLU-based linear ViTs.
As shown, \textit{Random} and \textit{Copy} achieve comparable accuracy across datasets.
Compared to both, \textit{MLPCopy} rises far above on all datasets; while \textit{FullCopy} yields further consistent gains over \textit{MLPCopy} by additionally copying the attention.\footnote{We focus only on the effect of loading the attention and MLP weights, 
as the ablation in~\Cref{supp:cls_ablation} shows that copying the remaining components yields no significant gain.
}

Therefore, the MLP blocks are \textit{operator-agnostic}: simply copying them already carries most of the benefit of the pre-trained weights, consistent with 
the view of MLPs as key-value memories that store learned knowledge~\citep{geva2021transformer}.
Compared with the attention findings above for Softmax-to-linear transfer, this indicates that the learned representation carried in MLP weights can be effectively transferred generally across various transfer settings, whereas the copied attention weights bring no reliable benefit on their own once the operator changes, similar to the architectural-mismatch findings in~\citet{qin2026attention}.
We conclude that what transfers effectively across the Softmax-to-linear boundary, by direct copying, is primarily the representation in the MLP blocks.

\subsection{The Recipe}
\label{subsec:recipe}

Having identified what is worth copying and what is worth distilling, we next ask whether the two transfer methods can be paired to reach the Softmax recovery target more effectively.
We answer by revisiting the distillation results in~\Cref{fig:distill_rescue} and the copying results in~\Cref{fig:factorial}, in both cases replacing the randomly initialized components with their effectively transferred counterparts.

Specifically, although distilled attention brings significant gains in~\Cref{fig:distill_rescue}, recovery to the Softmax target is still not consistent across datasets and linear operators.
Combining the copied MLPs can provide consistent further gains wherever the target is not yet reached, largely closing the remaining gap across operators and datasets.
In addition, although all four weight-initialization settings shown in~\Cref{fig:factorial} yield accuracy below the Softmax target, distilling the attention lifts every one of them, bringing \textit{MLPCopy} and \textit{FullCopy} to match the Softmax level.
Interestingly, \textit{MLPCopy w/ Distill} lands close to \textit{FullCopy w/ Distill}, showing that once the token routing behavior is distilled, the copied attention weights become nearly redundant, with no consistent advantage compared to copying-only settings.
Moreover, distillation also reverses the standing of the copied attention itself: 
while \textit{Copy} alone stays near the \textit{Random}-initialization tier, \textit{Copy w/ Distill} consistently exceeds \textit{Random w/ Distill}; we attribute this to the copied attention weights offering a useful warm start for recovering the routing behavior, although this advantage is subsumed once the MLP weights are copied.

Together, copying the operator-agnostic MLPs with the distilled attention can close the remaining gap and eventually match the Softmax recovery target, in some cases even surpassing it.
Therefore, we suggest one simple principle for reusing the pre-trained weights across the Softmax-to-linear boundary: copy what stays the same and distill what differs.
We provide further discussion on why copying the MLPs and distilling the attention outputs are complementary in~\Cref{supp:transfer_explanation}.

\begin{figure}[t]
    \centering
    \includegraphics[width=\linewidth]{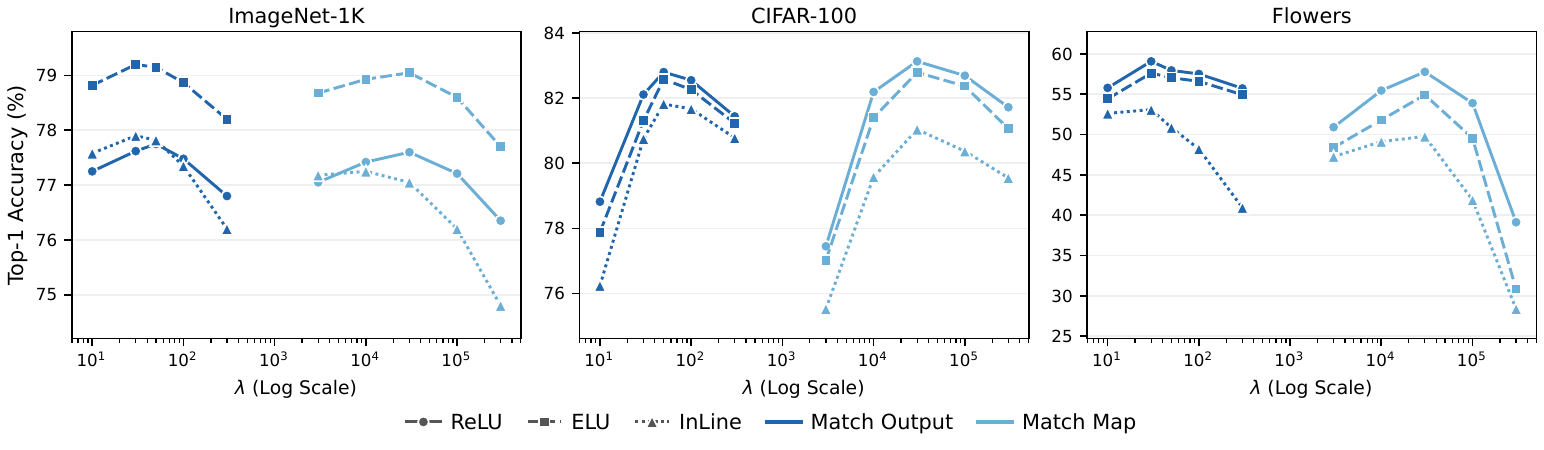}
    \vspace{-18pt}
    \caption{\textbf{Loss ablation on matching attention outputs \vs attention maps.}
    For \textit{Copy}-initialized ViT-Small students, we compare MSE matching on \textcolor[HTML]{2166ac}{attention \emph{outputs}} and \textcolor[HTML]{6baed6}{attention \emph{maps}} with varying $\lambda$ separately across datasets of different scales.
    Both matching targets show consistent accuracy trends under the $\lambda$-sweep.
    For ReLU and ELU, whose maps are proper distributions, the calibrated peaks are comparable.
    In contrast, for InLine, whose map is signed, map matching falls behind, suggesting that attention output matching is a more general and proper design.
    }
    \vspace{-12pt}
    \label{fig:loss_ablation}
\end{figure}

\subsection{Discussion and Analysis}
\label{subsec:loss_analysis}

\paragraph{Ablation on Distillation Loss Design.}
We further examine the loss-placement choice in~\Cref{subsec:distill} by sweeping the distillation loss weight $\lambda$ separately for output matching and map matching, each around the $\lambda$ with the peak accuracy, across datasets of different scales.
As shown in~\Cref{fig:loss_ablation}, for ReLU and ELU, whose attention maps are proper distributions, map matching and output matching yield comparable peak recovery, confirming that the recovery is robust to the specific loss form.
However, for InLine, whose map is a signed quasi-distribution due to its subtraction normalization, map matching falls consistently behind the output matching.
Therefore, a more general design for Softmax-to-linear attention distillation should match the routing behavior at the operator-agnostic attention output rather than the operator-specific attention maps.
%

\paragraph{Ablation on Distillation Loss Weight $\lambda$.}
Beyond the loss choice, the sweep in~\Cref{fig:loss_ablation} also shows that the output-matching peak is stable in $\lambda$: accuracy varies within a broad plateau around the peak, which roughly aligns across operators and datasets; we thus adopt $\lambda=50$ as the default distillation loss weight in our experiments.

\paragraph{Ablation on More Linear ViTs, Sizes, and Tasks.}
While all results in~\Cref{sec:localize} are with ReLU-based ViT-Small on classification, we verify the generalizability of these findings with comprehensive evaluations on various linear ViTs, more model sizes, and dense prediction tasks under the same experimental protocols.
As shown in~\Cref{supp:more_operators,supp:more_sizes,supp:imagenet_dense}, patterns consistent with those in~\Cref{fig:factorial} can also be observed, confirming the robustness of the transfer recipe across model types, sizes, and tasks.


\section{Conclusion}
\label{sec:conclusion}

In this work, we study whether linear ViTs can be initialized by reusing the pre-trained Softmax weights, and find that the most direct strategy fails: the attention weights are \textit{operator-specific}, and copying them can even fall below random initialization.
Instead, the attention's token routing behavior can be recovered through distillation with a proper loss design that matches the attention outputs.
The MLP weights, in contrast, are \textit{operator-agnostic}: simply copying them already carries most of the benefit of the pre-trained weights.
Paired together, the two close the remaining gap to the Softmax target and can even surpass it.
This yields a simple but general recipe: copy what stays the same and distill what differs.
We hope these findings sharpen the understanding of reusing pre-trained weights across attention operators and encourage further study of efficient transfer beyond the Softmax setting.

\section*{AI use statement}

We certify the use of AI to aid in the writing, polishing, and figure plotting of this manuscript, and the use of AI to aid code implementations of involved experiments in this work.
All AI-assisted content shown in this manuscript has been reviewed and verified by the authors.
We take responsibility for the final content of this work, including text, claims or artifacts produced with the aid of AI.

\section*{Ethics statement}

This work is primarily a methodological study of whether linear ViTs can benefit from pre-trained Softmax weights for initialization.
Its direct positive impact is to encourage more efficient and effective linear ViT initialization, and to help the community sharpen the understanding of reusing pre-trained weights across attention operators.
A potential negative impact is misinterpretation: our findings should not be read as ``Attention Transfer does not work across attention operators'', 
but rather as a recipe for how it should be conducted: copy what stays the same and distill what differs.
Finally, this paper relies on existing public pre-trained Softmax weights, some of which are trained on large-scale curated or web-scale datasets and may inherit dataset biases.
The proposed recipe for Softmax-to-linear transfer can make this transfer more effective, but it does not remove biases or safety concerns inherited from the original teacher weights.

\section*{Reproducibility statement}

To ensure the reproducibility of our results, we commit to making our source code publicly available upon publication. 
The code will include our implementations and scripts to replicate the empirical results presented in this manuscript. 
Comprehensive experimental setups of our analysis, including model zoo, transfer protocol, evaluation protocol, and selection of reference baselines,
are provided in~\Cref{subsec:exp_setup} and further detailed in~\Cref{supp:impl}.
Additionally, all experiments are repeated over 3 random seeds to ensure statistical significance.
\cref{tab:supp_seedvar} provides the statistical significance analysis of our obtained results.



{
\bibliographystyle{meta/iclr_2027.bst}
\bibliography{sections/11_references}
}

\newpage
\appendix
\renewcommand{\thesection}{\Alph{section}}
\renewcommand{\thetable}{\Alph{table}}
\renewcommand{\thefigure}{\Alph{figure}}
\renewcommand{\theequation}{\Alph{equation}}

\setcounter{section}{0}
\setcounter{table}{0}
\setcounter{figure}{0}
\setcounter{equation}{0}

\tableofcontents
\newpage

\section{Scope and Relation to Similar Methods}
\label{supp:scope}

We provide further discussion on the scope of this study and clarify its relation to existing methods for Softmax-to-linear transfer below.

\paragraph{Relation to linearization methods.}
Generally, our work is a controlled empirical study of \emph{initialization} of linear ViTs: we explore the reuse of pre-trained Softmax weights by factorizing them at the component level, test whether each component can be transferred as weights or as behavior, and validate our findings under fixed training recipes on various downstream tasks across five linear attention operators.
We do not claim to propose a new linearization method, and we do not claim to improve upon the methods below on their own benchmarks:

\begin{itemize}

    \item \textbf{LoLCATs}~\citep{zhang2025lolcats} linearizes large language models by training linear attention to match Softmax attention outputs under an MSE loss, followed by low-rank adaptation.
    The attention-output-matching objective used in~\Cref{subsec:distill} follows this established practice.
    Our contribution is not the loss itself, but the finding that this behavioral target is well-defined for ViTs and effective across all five linear operators, including those that never materialize an attention map.

    \item \textbf{ViT-AdaLA}~\citep{li2026vit} adapts pre-trained Softmax ViTs to vanilla linear attention through a dedicated pipeline of block-level attention alignment, final-layer feature alignment, and supervised fine-tuning, establishing a vision-specific use of attention alignment for one operator.
    Our findings instead offer a simple and operator-general recipe that requires no dedicated adaptation stage.

    \item \textbf{LiT}~\citep{wang2025lit} converts pre-trained diffusion transformers into linear ones, where its practical guideline is to load all parameters \emph{except} those of the linear attention and combine the selective inheritance with hybrid distillation on the predicted noise and variance.
    Our finding that MLP weights are operator-agnostic while attention weights are operator-specific is consistent with this guideline, and extends the observation from one diffusion setting to a component-level comparison across five linear operators, with the distillation signal placed on attention outputs rather than task predictions.

    \item \textbf{T$^5$}~\citep{li2026linearizing} converts pre-trained ViTs into the TTT architecture by inheriting projection, MLP, and normalization weights while randomly initializing the TTT inner operator, which corresponds to the \textit{FullCopy} setting in our work.
    With detailed attention-only and MLP-only ablations, our decomposition analysis instead identifies which component carries the benefit, and shows that the attention is better transferred by distillation than by copying.
    Our TTT student is the standard ViT$^3$ block of~\citet{han2026vit3} without T$^5$'s architectural modifications, which concerns the standard form of this architecture family.

    \item \textbf{DiD}~\citep{qin2026did} converts the Softmax-attention backbone of a trained detector into a linear-attention one without labels, keeping the downstream detector fixed and distilling the detector-facing interface tensors so that the converted backbone reproduces the features the detector expects.
    It addresses a different setting from ours: a post-hoc, label-free Softmax-to-linear conversion under a frozen detector, whereas we study how to initialize a linear ViT that is subsequently trained with supervision, and our dense-prediction results in~\Cref{supp:imagenet_dense} fine-tune the whole detector with a linear backbone transferred from Softmax pre-trained weights.
    Our findings are complementary: DiD shows that behavior-level alignment suffices to swap the attention operator with a fixed detector, while our decomposition identifies which and how pre-trained components should be copied or distilled at the initialization stage, when downstream heads are trainable.
\end{itemize}

\paragraph{Relation to Attention Transfer.}
Attention Transfer~\citep{li2024attention,qin2026attention} in vision studies whether transferring only the attention from a pre-trained Softmax ViT to a Softmax student with the same architecture can recover the original model's performance.
These studies show that Attention Copy, which keeps the teacher's attention fixed throughout student training by injecting the teacher's attention maps in every forward pass, is one effective way for such transfer.
Our \textit{Copy} setting is inspired by this, but provides a different operation: the copied pre-trained attention projection weights serve only as an initialization, and every parameter, including the copied ones, is trained jointly on downstream tasks.
This difference is deliberate due to the different focus on research questions: Attention Transfer asks whether the attention alone suffices to transfer a Softmax ViT, while we ask whether the Softmax attention \emph{weights} can initialize a linear attention operator.
The failure of \textit{Copy} in~\Cref{sec:main_results} is therefore a finding on weight-level initialization across the operator boundary, rather than a contradiction of Attention Transfer within the Softmax family.

\section{Additional Implementation Details}
\label{supp:impl}

We provide further implementation details to support reproducibility of our experimental setup below.

\subsection{Details of Model Zoo}
\label{supp:model_zoo}

In the Softmax-to-linear transfer experiments, all Softmax models follow the default implementation of the DeiT architecture~\citep{touvron2021training} across the Tiny, Small, and Base sizes.
Linear models differ from the Softmax ones only in the attention operator inside each transformer block, while all other components remain unchanged.
We study five representative linear ViT variants in this work, with detailed descriptions below:

\begin{itemize}
    \item \textbf{ELU and ReLU}~\citep{katharopoulos2020transformers} instantiate~\Cref{eq:linear_attn} with $\phi(x)=\mathrm{ELU}(x)+1$ and $\phi(x)=\mathrm{ReLU}(x)$, respectively.
    They are computed in the reordered linear-complexity form, with the projection weight shapes identical to those in Softmax.

    \item \textbf{InLine}~\citep{han2024bridging} uses the identity kernel with subtraction normalization to restore injectivity.
    With $d$ the per-head dimension, $s = d^{-1/2}$, and $\bar{k}$ the mean key, the map is $A_{ij} = \frac{s}{N}\, q_i^\top k_j + \frac{1}{N}\big(1 - s\, q_i^\top \bar{k}\big)$, whose rows sum to one while individual entries may be negative.
    It is computed with linear complexity as $\frac{s}{N}\, q_i^\top (K^\top V) + (1 - s\, q_i^\top \bar{k})\,\bar{v}$, where $\bar{v}$ is the mean value.
    We use the attention-only form without the local residual branch of the original block in their official implementation, which aggregates the $3\times3$ neighborhood of values with input-dependent weights, thus the projection weight shapes are identical to those in Softmax attention.

    \item \textbf{MHLA}~\citep{zhang2026mhla} partitions the token sequence into non-overlapping spatial blocks, treated as token-level heads, and applies linear attention within each block.
    Following the official implementation, we use the ReLU kernel $\phi(x)=\mathrm{ReLU}(x)$, with four $7\times7$ blocks on the $14\times14$ token grid.
    The per-block key-value summaries and normalizers are then mixed across blocks by a learned $1\times 1$ convolution mixing matrix, and, following the official implementation, a $5\times5$ depthwise convolution on the values provides a positional term that is added to the output before the output projection.
    All projection weights have the same shapes as in Softmax attention, with the mixing matrix and the depthwise convolution as additional components.
    
    \item \textbf{TTT} follows the ViT$^3$ block~\citep{han2026vit3}, which replaces the attention in ViTs with two inner models fitted on the fly for each image: a SwiGLU inner module with per-head weights $W_1, W_2 \in \mathbb{R}^{d\times d}$ and a $3\times3$ depthwise-convolution inner module $W_3$ as one additional head.
    Each inner model is updated by one closed-form gradient step on the key-value pairs of the current image and then applied to the queries.
    The two branch outputs are concatenated and mapped back to the model dimension $C$ by a $(C + d)\to C$ output projection.
    The $QKV$ projection therefore has the $3C$ rows of standard queries, keys, and values for the SwiGLU branch plus $3d$ rows for the convolutional branch; the inner weights, the extra rows, and the wider output projection are additional components.
\end{itemize}

\begin{table}[t]
   \centering
\small
\caption{\textbf{Implementation details of transfer protocol.}
We list the weight-initialization details of each component in the Softmax-to-linear transfer experiments.
Note that we follow the default implementations in MHLA and TTT to use sinusoidal positional embeddings and average pooling with no class token.
All other settings keep the class token and learned positional embeddings.
Mimetic and Impulse assume the standard $QK^\top$ attention parameterization and are therefore incompatible with TTT.
}
\setlength{\tabcolsep}{3pt}
\ra{1.02}
\footnotesize
\begin{tabular}{l l c c}
\textbf{Setting} & \textbf{Copied components} & \textbf{Class token} & \textbf{Pos. embedding} \\
\midrule
Random & -- & random & random \\
Mimetic & -- & random & random \\
Impulse & -- & random & random \\
\midrule
Copy & Attention & random & random \\
MLPCopy & MLPs & random & random \\
FullCopy & all Softmax weights & copied & copied \\
\midrule
Random w/ Distill & -- & random & random \\
Copy w/ Distill & Attention & random & random \\
MLPCopy w/ Distill & MLPs & random & random \\
FullCopy w/ Distill & all Softmax weights & copied & copied \\
\midrule
Softmax & all Softmax weights & copied & copied  \\
\end{tabular}
\label{tab:supp_protocol}

\end{table}

\subsection{Details of Transfer Protocol}
\label{supp:transfer_protocol}

Unless specifically mentioned, all reported results use the transfer source of the official ImageNet-1K DeiT pre-trained weights~\citep{touvron2021training} with the same size as the student.
Downstream heads for different tasks are randomly initialized.
We summarize the main weight-initialization settings used in our evaluations below, and also provide more details in~\Cref{tab:supp_protocol}:
\begin{itemize}
    \item \textbf{Copy} loads the pre-trained Softmax attention projection weights, including queries $Q$, keys $K$, values $V$, into every block of the linear ViT, while leaving all other components randomly initialized.
    For TTT, whose output projection has a different shape, only the rows of the standard $QKV$ are loaded.
    
    \item \textbf{MLPCopy} directly copies the MLP blocks from the pre-trained Softmax weights, while leaving all other components randomly initialized.
    
    \item \textbf{FullCopy}\footnote{Compared with \textit{Copy} or \textit{MLPCopy}, \textit{FullCopy} additionally loads the weights of the remaining components in ViTs, mainly the patch embedding, positional embedding, and class token. 
    We find that weight copying for these components does not yield significant performance gains, as shown in a small component-level ablation in~\Cref{supp:cls_ablation}.
    We therefore focus on the effect of initializing the attention and MLP weights.} copies all pre-trained Softmax weights.
    Unmatched components which are not in the default Softmax implementation, the two convolutions of MHLA and the inner weights, extra projection rows, and wider output projection of TTT, adopt the default initializations in their corresponding settings.
\end{itemize}
After initialization, every parameter of each setting is trained jointly on downstream tasks, and none of the student's weights are frozen.

\subsection{Details of Evaluation Protocol}
\label{supp:eval_protocol}

\Cref{tab:supp_datasets} summarizes all the datasets used in our Softmax-to-linear transfer experiments.
We use the standard train/test splits and report the Top-1 Accuracy of each dataset.
All reported results are averaged over 3 random seeds to ensure statistical significance, with the corresponding analysis provided in~\Cref{supp:significance}.
\Cref{tab:supp_recipe} summarizes the training recipes used in our experiments, adapted from~\citet{xu2024initializing}, with training schedules of 300 or 600 epochs depending on the dataset scale.
All experiments are conducted on NVIDIA A100 GPUs.

\begin{table}[t]
\centering
\small
\caption{\textbf{Details of datasets and training schedules.}}
\vspace{-3pt}
\label{tab:supp_datasets}
\ra{1.02}
\footnotesize
\setlength{\tabcolsep}{3pt}
\begin{tabular}{lrrrrr}
\textbf{Dataset} & \textbf{Classes} & \textbf{Train} & \textbf{Test / Val} & \textbf{Epochs} & \textbf{Warm-up} \\
\midrule
ImageNet-1K~\citep{deng2009imagenet} & 1,000 & 1,281,167 & 50,000 & 300 & 50 \\
CIFAR-10~\citep{krizhevsky2009learning} & 10 & 50,000 & 10,000 & 300 & 50 \\
CIFAR-100~\citep{krizhevsky2009learning} & 100 & 50,000 & 10,000 & 300 & 50 \\
STL-10~\citep{coates2011analysis} & 10 & 5,000 & 8,000 & 300 & 50 \\
Food~\citep{bossard2014food} & 101 & 75,750 & 25,250 & 300 & 50 \\
Flowers~\citep{nilsback2008automated} & 102 & 1,020 & 6,149 & 600 & 100 \\
Pets~\citep{parkhi2012cats} & 37 & 3,680 & 3,669 & 600 & 100 \\
\end{tabular}
\vspace{-6pt}
\end{table}

\begin{table}[t]
   \centering
\small
\caption{\textbf{Training recipes for transfer datasets and ImageNet-1K.}
}
\vspace{-3pt}
\ra{1.02}
\setlength{\tabcolsep}{2pt}
\footnotesize
\begin{tabular}{l l l}
    \textbf{Config} & \textbf{Transfer datasets} & \textbf{ImageNet-1K} \\
    \midrule
    \textbf{Optimizer} & AdamW & AdamW \\
    \textbf{Base Learning Rate} & 2e-3 & 2e-3 \\
    \textbf{Weight Decay} & 0.05 & 0.05 \\
    \textbf{Optimizer Momentum} & $\beta_1=0.9$, $\beta_2=0.999$ & $\beta_1=0.9$, $\beta_2=0.999$ \\
    \textbf{Layer-wise LR Decay} & -- & -- \\
    \textbf{Batch Size} & 512 & 4096 \\
    \textbf{LR Schedule} & cosine (min.\ 1e-6) & cosine (min.\ 1e-6) \\
    \textbf{Gradient Clipping} & 3.0 & 3.0 \\
    \textbf{Augmentation} & RRC, flip, color jitter 0.4 & RRC, flip, color jitter 0.4 \\
    \textbf{Label Smoothing} & 0.1 & 0.1 \\
    \textbf{Drop Path} & 0 & 0 (Tiny, Small); 0.1 (Base) \\
    \textbf{Input / Eval.\ Crop} & $224^2$ / center crop 0.9 & $224^2$ / center crop 0.9 \\
    \textbf{Distillation Loss} & attention-output MSE & attention-output MSE \\
    \textbf{Distillation Loss Weight $\lambda$} & 50 & 50 \\
    \textbf{Random Seeds} & 3 & 3 \\
\end{tabular}
\label{tab:supp_recipe}

   \vspace{-9pt}
\end{table}

\subsection{Details of Attention Distillation}
\label{supp:distill_details}

The \textit{w/ Distill} settings follow the Attention Distillation used in~\citet{li2024attention,qin2026attention}, with a frozen pre-trained Softmax teacher and a per-block distillation loss, but place the loss on the attention outputs instead of the attention maps, as discussed in~\Cref{subsec:distill}.
The training aims to match the student's attention outputs to the teacher’s through an auxiliary MSE objective at every block during the downstream fine-tuning, leading to the final training objective as $\mathcal{L} = \mathcal{L}_{\mathrm{task}} + \lambda\, \mathcal{L}_{\mathrm{distill}}$.
Specifically, for MHLA and TTT, which have no class token, only the patch tokens are matched when computing the loss.
As discussed in~\Cref{subsec:loss_analysis}, we use $\lambda = 50$ as the default distillation loss weight across all operators, datasets, and initialization settings.

For the loss ablation in~\Cref{fig:loss_ablation}, the map-matching loss is computed with the recorded inputs of the attention module in each block.
The teacher map is the Softmax map, while the student map is the row-normalized kernel map for ReLU and ELU, and the signed subtraction-normalized map for InLine.
The implementations of MHLA and TTT never materialize an attention map, thus are excluded from this ablation.
Inspired by~\citet{qin2026attention}, since the two losses differ in magnitude, each is swept over its own range: $\lambda \in \{10, 30, 50, 100, 300\}$ for output matching and $\lambda \in \{3\times10^{3}, 10^{4}, 3\times10^{4}, 10^{5}, 3\times10^{5}\}$ for map matching, on the three reported datasets.

\section{Additional Results and Analysis}
\label{supp:results}
We provide additional experimental results and analysis to further support the generalizability and robustness of our findings below.

\subsection{Additional Results on More Linear Operators}
\label{supp:more_operators}

\begin{figure}[ht]
    \centering
    \includegraphics[width=\linewidth]{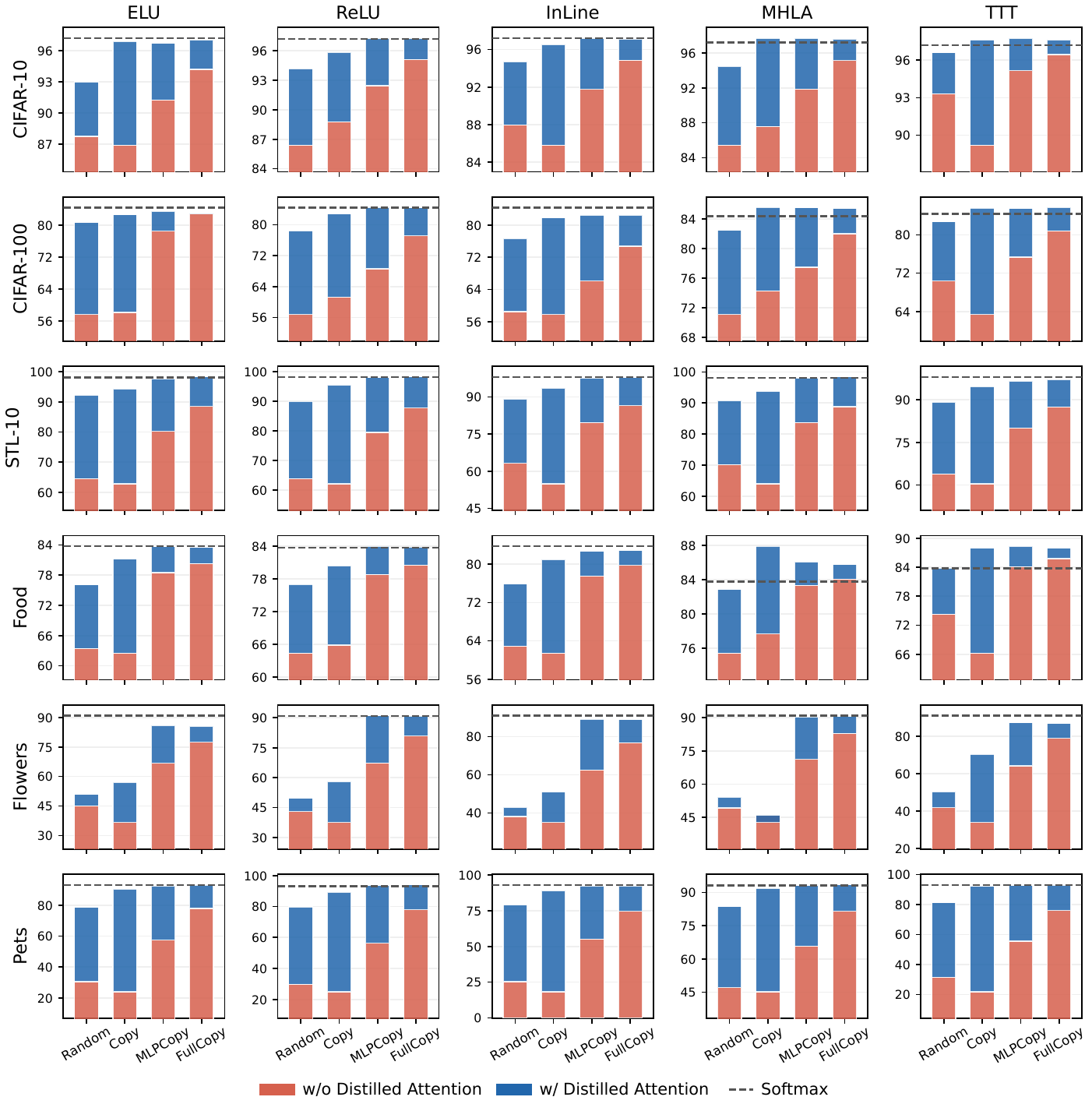}
    \vspace{-21pt}
    \caption{\textbf{Extended comparison results across linear operators at Small size.}
    Each linear ViT is initialized under four weight-initialization settings: \textit{Random}, \textit{Copy}, \textit{MLPCopy}, and \textit{FullCopy}; each w/ and w/o attention distillation.
    Results are reported with the Top-1 Accuracy (\%) for all five linear ViTs (columns) on the six transfer datasets (rows) at Small size.
    In each panel, the \textcolor[HTML]{d6604d}{red} bar is the accuracy of the copy setting alone, the \textcolor[HTML]{2166ac}{blue} block is the additional gain from attention distillation, and the dashed line is the fine-tuned Softmax target.
    }
    \label{fig:factorial_ops}
    \vspace{-12pt}
\end{figure}

\Cref{fig:factorial} reports the comparison of the four weight-initialization settings with the ReLU-based linear ViTs.
We provide comprehensive results for all five linear operators in~\Cref{fig:factorial_ops} under the same protocol.
As shown, the patterns revealed in~\Cref{sec:localize} hold consistently for every operator:
\textit{Copy} remains near \textit{Random}, while \textit{MLPCopy} alone lifts every operator far above \textit{Random} on every dataset and \textit{FullCopy} yields further consistent gains.
Adding the distilled attention additionally lifts the accuracy in nearly every setting:
\textit{Copy w/ Distill} exceeds \textit{Random w/ Distill} for almost every operator and dataset, confirming the warm-start effect of the copied attention weights across operators, as discussed in~\Cref{subsec:recipe}.
\textit{MLPCopy w/ Distill} lands close to \textit{FullCopy w/ Distill} on every operator and dataset, with the final recipe eventually matching or even surpassing the Softmax recovery target on most datasets.

\subsection{Additional Results on More Model Sizes}
\label{supp:more_sizes}

\begin{figure}[ht]
    \centering
    \includegraphics[width=\linewidth]{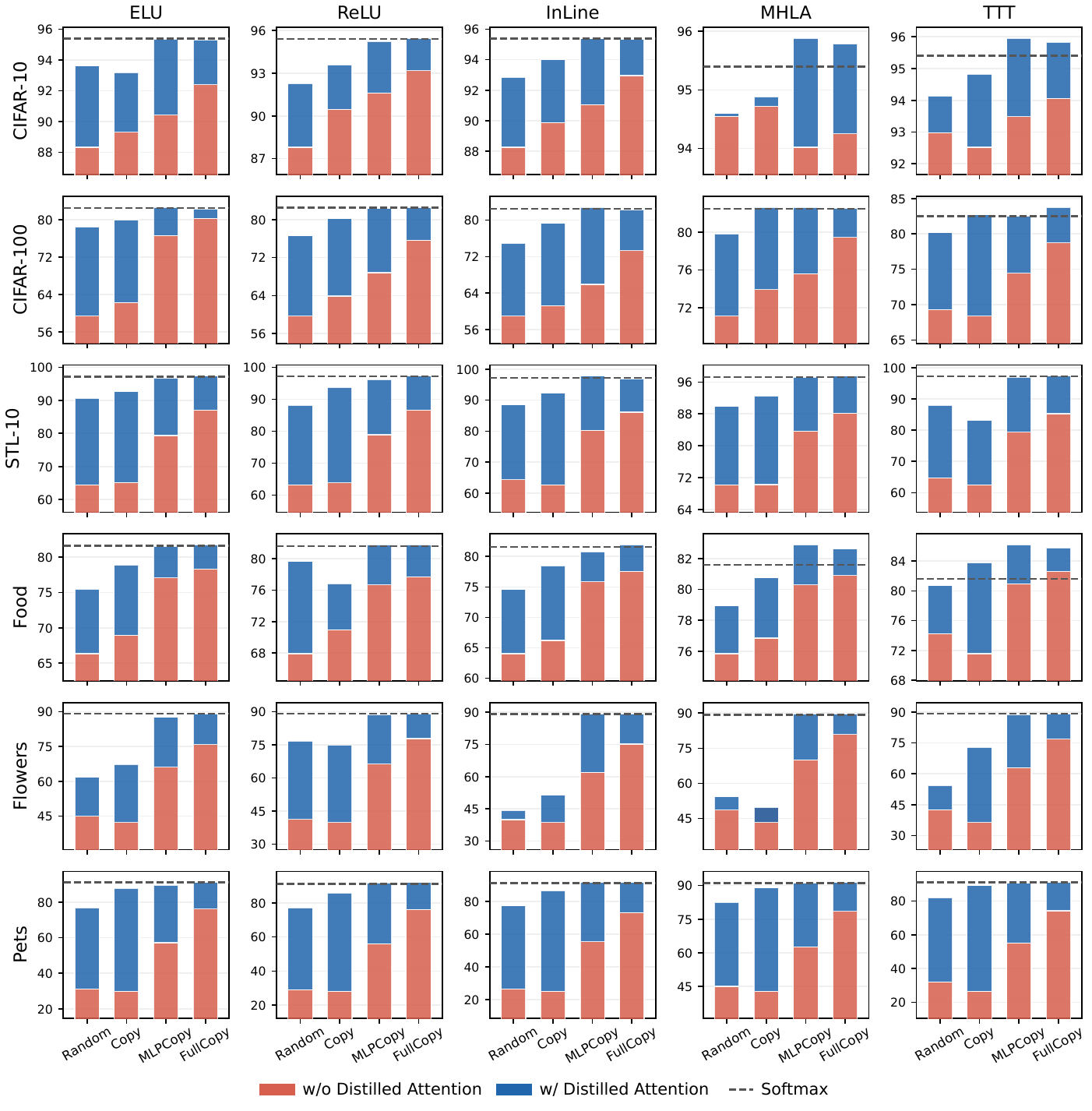}
    \vspace{-21pt}
    \caption{\textbf{Extended comparison results across linear operators at Tiny size.}
    Each linear ViT is initialized under four weight-initialization settings: \textit{Random}, \textit{Copy}, \textit{MLPCopy}, and \textit{FullCopy}; each w/ and w/o attention distillation.
    Results are reported with the Top-1 Accuracy (\%) for all five linear ViTs (columns) on the six transfer datasets (rows) at Tiny size.
    In each panel, the \textcolor[HTML]{d6604d}{red} bar is the accuracy of the copy setting alone, the \textcolor[HTML]{2166ac}{blue} block is the additional gain from attention distillation, and the dashed line is the fine-tuned Softmax target.
    }
    \label{fig:factorial_ops_tiny}
    \vspace{-12pt}
\end{figure}

To show the generalizability of our findings across model sizes, we repeat the comparison in~\Cref{fig:factorial_ops} for all five linear operators at Tiny size under the same protocol in~\Cref{fig:factorial_ops_tiny}.
We further extend the evaluation to Base size on ImageNet-1K and dense prediction tasks in~\Cref{supp:imagenet_dense}.
As shown, the patterns revealed in~\Cref{sec:localize} hold consistently for every operator at the smaller size as well, with smaller margins: \textit{FullCopy} improves over \textit{MLPCopy} on every dataset, distillation lifts nearly every setting, and \textit{MLPCopy w/ Distill} lands close to \textit{FullCopy w/ Distill}, which matches the Softmax target.

\subsection{Additional Results on ImageNet and Dense Prediction}
\label{supp:imagenet_dense}

As mentioned, we extend the component-level comparison of four weight-initialization settings to ImageNet-1K at Base size and further test the resulting transferred weights as backbones for dense prediction tasks including object detection, instance segmentation, and semantic segmentation.
We report results with ReLU and TTT, by training Base models on ImageNet-1K from each of the four weight-initialization settings, w/ and w/o attention distillation from the frozen DeiT-B teacher, resulting in eight transferred linear model weights per operator.
Each linear model is then fine-tuned with Mask R-CNN~\citep{he2017mask} on COCO~\citep{lin2014microsoft} under the 1$\times$ schedule, or with UperNet~\citep{xiao2018unified} on ADE20K~\citep{zhou2017scene} for 160k iterations, following the plain-ViT baseline implementations in~\citet{chen2023vit}.

\Cref{tab:supp_imagenet_dense_relu,tab:supp_imagenet_dense_ttt} report the results for ReLU and TTT, respectively.
As shown, evaluations on ImageNet-1K and on the dense tasks align consistently with the findings on the six classification datasets in our discussion:
\textit{Copy} brings no reliable advantage over \textit{Random}, while \textit{MLPCopy} carries most of the benefit of the pre-trained weights and \textit{FullCopy} further adds a small additional gain, confirming that the attention weights are operator-specific and the MLP weights are operator-agnostic beyond classification.
Attention distillation lifts every initialization effectively, with \textit{MLPCopy w/ Distill} generally landing close to \textit{FullCopy w/ Distill}, which closes most of the gap to the Softmax target and matches or even exceeds it.

\begin{table}[t]
   \centering
\small
\caption{\textbf{ImageNet-1K and dense prediction results with ReLU at Base size.}
All settings adopt transferred weights from DeiT and are then fine-tuned with Mask R-CNN (1$\times$) on COCO and UperNet (160k iters) on ADE20K.
We report the Top-1 Accuracy (\%) on ImageNet-1K, box and mask AP on COCO, and mIoU with single and multi-scale (+MS) on ADE20K, respectively.
w/ Distill with $\checkmark$ indicates applying attention distillation during the Softmax-to-linear transfer, and colored deltas indicate gains relative to the same setting w/o distillation.
\colorbox[HTML]{d8d8d8}{Softmax}~row refers to the Softmax reference and the plain-ViT baselines of~\citet{chen2023vit}.
}
\vspace{-2pt}
\label{tab:supp_imagenet_dense_relu}
\ra{1.02}
\setlength{\tabcolsep}{2.2pt}
\footnotesize
\resizebox{\linewidth}{!}{%
\begin{tabular}{l c c cccccc cc}
& w/ & IN-1K & \multicolumn{6}{c}{COCO} & \multicolumn{2}{c}{ADE20K} \\
\cmidrule(lr){3-3} \cmidrule(lr){4-9} \cmidrule(lr){10-11}
& Distill & Acc & AP$^b$ & AP$^b_{50}$ & AP$^b_{75}$ & AP$^m$ & AP$^m_{50}$ & AP$^m_{75}$ & mIoU & +MS \\
\midrule
\cellcolor[HTML]{d8d8d8}Softmax & \cellcolor[HTML]{d8d8d8} & \cellcolor[HTML]{d8d8d8}81.8 & \cellcolor[HTML]{d8d8d8}42.9 & \cellcolor[HTML]{d8d8d8}65.7 & \cellcolor[HTML]{d8d8d8}46.8 & \cellcolor[HTML]{d8d8d8}39.4 & \cellcolor[HTML]{d8d8d8}62.6 & \cellcolor[HTML]{d8d8d8}42.0 & \cellcolor[HTML]{d8d8d8}46.1 & \cellcolor[HTML]{d8d8d8}47.1 \\
\addlinespace[3pt]
Random & & 76.5 & 34.0 & 56.5 & 37.6 & 32.1 & 55.5 & 34.4 & 34.9 & 35.7 \\
\addlinespace[1.2pt]
& $\checkmark$ & 79.4 {\scriptsize\color[HTML]{2166ac}+2.9} & 36.2 {\scriptsize\color[HTML]{2166ac}+2.2} & 59.0 {\scriptsize\color[HTML]{2166ac}+2.5} & 40.3 {\scriptsize\color[HTML]{2166ac}+2.7} & 35.8 {\scriptsize\color[HTML]{2166ac}+3.7} & 59.2 {\scriptsize\color[HTML]{2166ac}+3.7} & 37.9 {\scriptsize\color[HTML]{2166ac}+3.5} & 40.7 {\scriptsize\color[HTML]{2166ac}+5.8} & 41.8 {\scriptsize\color[HTML]{2166ac}+6.1} \\
\addlinespace[3pt]
Copy & & 75.6 & 33.3 & 55.9 & 37.1 & 31.6 & 55.0 & 33.7 & 35.2 & 35.8 \\
\addlinespace[1.2pt]
& $\checkmark$ & 79.9 {\scriptsize\color[HTML]{2166ac}+4.3} & 36.9 {\scriptsize\color[HTML]{2166ac}+3.6} & 59.3 {\scriptsize\color[HTML]{2166ac}+3.4} & 40.7 {\scriptsize\color[HTML]{2166ac}+3.6} & 36.0 {\scriptsize\color[HTML]{2166ac}+4.4} & 59.6 {\scriptsize\color[HTML]{2166ac}+4.6} & 38.2 {\scriptsize\color[HTML]{2166ac}+4.5} & 41.5 {\scriptsize\color[HTML]{2166ac}+6.3} & 42.1 {\scriptsize\color[HTML]{2166ac}+6.3} \\
\addlinespace[3pt]
MLPCopy & & 80.2 & 41.4 & 63.9 & 44.9 & 36.5 & 60.1 & 38.7 & 43.5 & 44.4 \\
\addlinespace[1.2pt]
& $\checkmark$ & 81.4 {\scriptsize\color[HTML]{2166ac}+1.2} & 42.2 {\scriptsize\color[HTML]{2166ac}+0.8} & 64.6 {\scriptsize\color[HTML]{2166ac}+0.7} & \textbf{45.9 {\scriptsize\color[HTML]{2166ac}+1.0}} & 38.2 {\scriptsize\color[HTML]{2166ac}+1.7} & 62.0 {\scriptsize\color[HTML]{2166ac}+1.9} & 40.7 {\scriptsize\color[HTML]{2166ac}+2.0} & 45.2 {\scriptsize\color[HTML]{2166ac}+1.7} & 45.9 {\scriptsize\color[HTML]{2166ac}+1.5} \\
\addlinespace[3pt]
FullCopy & & 80.8 & 41.5 & 64.1 & 45.2 & 36.7 & 60.4 & 39.2 & 43.9 & 44.6 \\
\addlinespace[1.2pt]
& $\checkmark$ & \textbf{82.0 {\scriptsize\color[HTML]{2166ac}+1.2}} & \textbf{42.4 {\scriptsize\color[HTML]{2166ac}+0.9}} & \textbf{64.9 {\scriptsize\color[HTML]{2166ac}+0.8}} & \textbf{45.9 {\scriptsize\color[HTML]{2166ac}+0.7}} & \textbf{39.0 {\scriptsize\color[HTML]{2166ac}+2.3}} & \textbf{62.4 {\scriptsize\color[HTML]{2166ac}+2.0}} & \textbf{41.3 {\scriptsize\color[HTML]{2166ac}+2.1}} & \textbf{45.3 {\scriptsize\color[HTML]{2166ac}+1.4}} & \textbf{46.4 {\scriptsize\color[HTML]{2166ac}+1.8}} \\
\end{tabular}}
\vspace{3pt}

   \vspace{6pt}
   \centering
\small
\caption{\textbf{ImageNet-1K and dense prediction results with TTT at Base size.}
All settings adopt transferred weights from DeiT and are then fine-tuned with Mask R-CNN (1$\times$) on COCO and UperNet (160k iters) on ADE20K.
We report the Top-1 Accuracy (\%) on ImageNet-1K, box and mask AP on COCO, and mIoU with single and multi-scale (+MS) on ADE20K, respectively.
w/ Distill with $\checkmark$ indicates applying attention distillation during the Softmax-to-linear transfer, and colored deltas indicate gains relative to the same setting w/o distillation.
\colorbox[HTML]{d8d8d8}{Softmax}~row refers to the Softmax reference and the plain-ViT baselines of~\citet{chen2023vit}.
}
\vspace{-2pt}
\label{tab:supp_imagenet_dense_ttt}
\ra{1.02}
\setlength{\tabcolsep}{2.2pt}
\footnotesize
\resizebox{\linewidth}{!}{%
\begin{tabular}{l c c cccccc cc}
& w/ & IN-1K & \multicolumn{6}{c}{COCO} & \multicolumn{2}{c}{ADE20K} \\
\cmidrule(lr){3-3} \cmidrule(lr){4-9} \cmidrule(lr){10-11}
& Distill & Acc & AP$^b$ & AP$^b_{50}$ & AP$^b_{75}$ & AP$^m$ & AP$^m_{50}$ & AP$^m_{75}$ & mIoU & +MS \\
\midrule
\cellcolor[HTML]{d8d8d8}Softmax & \cellcolor[HTML]{d8d8d8} & \cellcolor[HTML]{d8d8d8}81.8 & \cellcolor[HTML]{d8d8d8}42.9 & \cellcolor[HTML]{d8d8d8}65.7 & \cellcolor[HTML]{d8d8d8}46.8 & \cellcolor[HTML]{d8d8d8}39.4 & \cellcolor[HTML]{d8d8d8}62.6 & \cellcolor[HTML]{d8d8d8}42.0 & \cellcolor[HTML]{d8d8d8}46.1 & \cellcolor[HTML]{d8d8d8}47.1 \\
\addlinespace[3pt]
Random & & 82.2 & 36.4 & 58.9 & 40.3 & 34.3 & 57.7 & 36.2 & 35.1 & 35.7 \\
\addlinespace[1.2pt]
& $\checkmark$ & 82.3 {\scriptsize\color[HTML]{2166ac}+0.1} & 40.5 {\scriptsize\color[HTML]{2166ac}+4.1} & 63.0 {\scriptsize\color[HTML]{2166ac}+4.1} & 44.2 {\scriptsize\color[HTML]{2166ac}+3.9} & 37.4 {\scriptsize\color[HTML]{2166ac}+3.1} & 60.9 {\scriptsize\color[HTML]{2166ac}+3.2} & 39.5 {\scriptsize\color[HTML]{2166ac}+3.3} & 44.5 {\scriptsize\color[HTML]{2166ac}+9.4} & 45.5 {\scriptsize\color[HTML]{2166ac}+9.8} \\
\addlinespace[3pt]
Copy & & 81.0 & 35.4 & 58.1 & 39.2 & 33.5 & 57.0 & 35.9 & 35.1 & 36.2 \\
\addlinespace[1.2pt]
& $\checkmark$ & 82.2 {\scriptsize\color[HTML]{2166ac}+1.2} & 41.0 {\scriptsize\color[HTML]{2166ac}+5.6} & 63.6 {\scriptsize\color[HTML]{2166ac}+5.5} & 44.7 {\scriptsize\color[HTML]{2166ac}+5.5} & 38.1 {\scriptsize\color[HTML]{2166ac}+4.6} & 61.9 {\scriptsize\color[HTML]{2166ac}+4.9} & 40.3 {\scriptsize\color[HTML]{2166ac}+4.4} & 45.2 {\scriptsize\color[HTML]{2166ac}+10.1} & 46.2 {\scriptsize\color[HTML]{2166ac}+10.0} \\
\addlinespace[3pt]
MLPCopy & & 82.9 & 42.9 & 65.2 & 46.5 & 38.0 & 61.5 & 40.1 & 44.3 & 45.3 \\
\addlinespace[1.2pt]
& $\checkmark$ & 83.0 {\scriptsize\color[HTML]{2166ac}+0.1} & 43.2 {\scriptsize\color[HTML]{2166ac}+0.3} & 65.7 {\scriptsize\color[HTML]{2166ac}+0.5} & 46.8 {\scriptsize\color[HTML]{2166ac}+0.3} & 39.1 {\scriptsize\color[HTML]{2166ac}+1.1} & 62.9 {\scriptsize\color[HTML]{2166ac}+1.4} & \textbf{41.5 {\scriptsize\color[HTML]{2166ac}+1.4}} & 46.7 {\scriptsize\color[HTML]{2166ac}+2.4} & 47.5 {\scriptsize\color[HTML]{2166ac}+2.2} \\
\addlinespace[3pt]
FullCopy & & 83.1 & 43.1 & 65.9 & 47.0 & 38.7 & 62.3 & 41.0 & 45.2 & 46.0 \\
\addlinespace[1.2pt]
& $\checkmark$ & \textbf{83.2 {\scriptsize\color[HTML]{2166ac}+0.1}} & \textbf{43.3 {\scriptsize\color[HTML]{2166ac}+0.2}} & \textbf{66.1 {\scriptsize\color[HTML]{2166ac}+0.2}} & \textbf{47.6 {\scriptsize\color[HTML]{2166ac}+0.6}} & \textbf{39.4 {\scriptsize\color[HTML]{2166ac}+0.7}} & \textbf{63.1 {\scriptsize\color[HTML]{2166ac}+0.8}} & \textbf{41.5 {\scriptsize\color[HTML]{2166ac}+0.5}} & \textbf{47.1 {\scriptsize\color[HTML]{2166ac}+1.9}} & \textbf{47.9 {\scriptsize\color[HTML]{2166ac}+1.9}} \\
\end{tabular}}
\vspace{3pt}

   \vspace{6pt}
   \centering
\small
\caption{\textbf{Statistical significance analysis on ImageNet-1K results.}
We report the standard deviation of the Top-1 Accuracy (\%) on ImageNet-1K results over 3 random seeds.
}
\vspace{-2pt}
\label{tab:supp_seedvar}
\setlength{\tabcolsep}{4pt}
\ra{1.02}
\footnotesize
\begin{tabular}{c cccc cccc}
 & \multicolumn{4}{c}{ReLU} & \multicolumn{4}{c}{TTT} \\
\cmidrule(lr){2-5} \cmidrule(lr){6-9}
 & Random & Copy & MLPCopy & FullCopy & Random & Copy & MLPCopy & FullCopy \\
\midrule
Acc. (w/o Distill) & 76.5 & 75.6 & 80.2 & 80.8 & 82.2 & 81.0 & 82.9 & 83.1 \\
$\pm$ std & 0.1 & 0.1 & 0.2 & 0.1 & 0.1 & 0.1 & 0.1 & 0.1 \\
\midrule
Acc. (w/ Distill) & 79.4 & 79.9 & 81.4 & 82.0 & 82.3 & 82.2 & 83.0 & 83.2 \\
$\pm$ std & 0.2 & 0.2 & 0.2 & 0.2 & 0.1 & 0.2 & 0.1 & 0.1 \\
\end{tabular}
\vspace{3pt}

   \vspace{-15pt}
\end{table}

\subsection{Additional Results on Copying the Remaining Components}
\label{supp:cls_ablation}

Compared with \textit{Copy} and \textit{MLPCopy}, \textit{FullCopy} additionally copies the weights of the remaining components beyond the attention and MLP blocks in ViTs.
These mainly include the patch embedding, positional embedding, and class token. 
One natural hypothesis is that the performance gain of \textit{FullCopy} also stems significantly from loading these components.
We therefore conduct a component-level ablation in~\Cref{tab:supp_cls_ablation} to isolate the contribution of loading these three components.
Specifically, we gradually remove the weight loading of these components from \textit{FullCopy}: first the class token (Variant 1), then the positional embedding (Variant 2), and finally the patch embedding (Variant 3), while keeping all attention and MLP weights copied.
The removed components are randomly initialized.
As shown, removing the weight copying for these components does not yield significant accuracy drops, compared with the substantial benefits from loading the attention and MLP weights.
Thus, we focus only on the effect of initializing the attention and MLP weights in our discussion.

\begin{table}[t]
\centering
\small
\caption{\textbf{Ablation on copying the remaining components.}
Starting from \textit{FullCopy}, we gradually remove the weight loading of the class token, positional embedding, and patch embedding, while all attention and MLP weights remain copied.
We report the Top-1 Accuracy (\%) on ImageNet-1K with ReLU at Base size, with colored deltas relative to the \textcolor{gray}{\textit{Random}} baseline.
}
\vspace{-2pt}
\label{tab:supp_cls_ablation}
\ra{1.02}
\setlength{\tabcolsep}{3pt}
\footnotesize
\begin{tabular}{l | >{\centering\arraybackslash}m{1.5cm}     >{\centering\arraybackslash}m{1.5cm}     >{\centering\arraybackslash}m{1.5cm}     >{\centering\arraybackslash}m{1.5cm}     >{\centering\arraybackslash}m{1.5cm}      | cc}
& \multicolumn{5}{c|}{\textbf{Weight Copying}} & \multicolumn{2}{c}{\textbf{Accuracy}} \\
& Attn. & MLP & Patch emb. & Pos. emb. & Cls. token & w/o Distill & w/ Distill \\
\midrule
FullCopy & \checkmark & \checkmark & \checkmark & \checkmark & \checkmark & 80.8 {\scriptsize\color[HTML]{2166ac}+4.3} & 82.0 {\scriptsize\color[HTML]{2166ac}+2.6} \\
Variant 1 & \checkmark & \checkmark & \checkmark & \checkmark &   & 80.7 {\scriptsize\color[HTML]{2166ac}+4.2} & 81.8 {\scriptsize\color[HTML]{2166ac}+2.4} \\
Variant 2 & \checkmark & \checkmark & \checkmark &  &   & 80.6 {\scriptsize\color[HTML]{2166ac}+4.1} & 81.9 {\scriptsize\color[HTML]{2166ac}+2.5} \\
Variant 3 & \checkmark & \checkmark &  &  &   & 80.7 {\scriptsize\color[HTML]{2166ac}+4.2} & 81.9 {\scriptsize\color[HTML]{2166ac}+2.5} \\
MLPCopy &  & \checkmark &  &  &   & 80.2 {\scriptsize\color[HTML]{2166ac}+3.7} & 81.4 {\scriptsize\color[HTML]{2166ac}+2.0} \\
Copy & \checkmark &  &  &  &   & 75.6 {\scriptsize\color[HTML]{d6604d}-0.9} & 79.9 {\scriptsize\color[HTML]{2166ac}+0.5} \\
\color{gray} Random & & &  &  &   & \color{gray} 76.5 & \color{gray} 79.4 \\
\end{tabular}
\end{table}

\subsection{Additional Results on Statistical Significance}
\label{supp:significance}

We further provide the statistical significance analysis of the ImageNet-1K results reported in~\Cref{tab:supp_imagenet_dense_relu,tab:supp_imagenet_dense_ttt}. 
We report the standard deviation of the Top-1 Accuracy under both w/ and w/o distillation across 3 random seeds in~\Cref{tab:supp_seedvar}.

\section{Additional Interpretation of Softmax-to-Linear Transfer}
\label{supp:transfer_explanation}

We provide a functional interpretation of our component-level findings in~\Cref{sec:main_results,sec:localize}: why the Softmax attention weights are operator-specific, why the MLP weights are operator-agnostic, and why copying the latter and distilling the former are complementary.
In the discussion below, we omit the block index $\ell$, the scaling factor $1/\sqrt{d_k}$, and the residual connection in MLPs in the formulation for simplicity.

\subsection{Preservation under Weight Copying}
\label{supp:functional_compatibility}

Pre-trained Softmax attention weights are optimized jointly with the Softmax computation operator during the pre-training stage.
For such fixed projection weights $Q$, $K$, and $V$, replacing the Softmax operator in~\Cref{eq:softmax_attn} with a linear-complexity alternative via a kernel operator as in~\Cref{eq:linear_attn} changes the output: $f_\text{linear} \neq f_\text{softmax}$ in general, as widely discussed in previous studies on linear attention~\citep{katharopoulos2020transformers,zhang2024hedgehog,han2024bridging}, because the Softmax map $\mathrm{softmax}(QK^\top)$ and the normalized kernel map
\begin{equation}
    A_\phi = \mathrm{diag}\big(\phi(Q)\phi(K)^\top\mathbf{1}_N\big)^{-1}\phi(Q)\phi(K)^\top,
    \label{eq:supp_kernel_map}
\end{equation}
are computed with different formulations although using the same $Q$ and $K$.
Simply copying the projection weights therefore does not preserve the pre-trained attention function, which is the functional meaning of the Softmax attention weights being \textit{operator-specific}.
In addition, when the remaining components are randomly initialized, as in \textit{Copy}, the inputs to the copied attention projections also differ from those seen during pre-training, leading to a further mismatch in the input features on top of the replaced operator.

MLPs have the opposite property:
the architecture remains unchanged across the Softmax-to-linear transfer, and the token-wise computation is fully determined by the parameters.
Copying the MLP weights therefore preserves this computation exactly for any input, which is the functional meaning of the MLP weights being \textit{operator-agnostic}.
What copying does not guarantee is that the student MLP receives the same input as the teacher MLP, but this input is affected by the output from the attention module in the same transformer block.

\subsection{Why Copying MLPs and Distilling Attention Are Complementary}
\label{supp:copy_distill_complementarity}

Let $z_s$ and $z_t$ denote the inputs to the corresponding MLPs of the linear student and the Softmax teacher, and let $M_s$ and $M_t$ denote the two corresponding MLP computations, respectively.
Their output difference then can be decomposed as:
\begin{equation}
    M_s(z_s) - M_t(z_t)
    = \underbrace{M_s(z_s) - M_t(z_s)}_{\text{computation mismatch}}
    + \underbrace{M_t(z_s) - M_t(z_t)}_{\text{input mismatch}}.
    \label{eq:supp_mlp_decomposition}
\end{equation}
At initialization, \textit{MLPCopy} removes the first term with the identical $M_s$ and $M_t$.
The second term remains, which is driven mainly by the attention operator replaced from Softmax to linear.
Attention distillation targets this term directly: by matching the outputs of the corresponding attention modules, it supervises this input mismatch and thus closes the gap through the auxiliary training objective.
The two transfers thus address the two terms of~\Cref{eq:supp_mlp_decomposition}: copying preserves the computation that stays the same, and distillation recovers the computation that the attention operator changes.

\section{Limitations}
\label{supp:limitations}

Our study is empirical and its scope is bounded in the following three ways.
First, all main Softmax-to-linear transfer experiments use ImageNet-1K DeiT pre-trained weights as the source, while other Softmax weights are only evaluated in the robustness check of~\Cref{subsec:reversal_robustness}.
Thus, the effectiveness of the final recipe is validated for one teacher family, although we expect it to hold across different pre-trained weights.
Second, the proposed recipe for Softmax-to-linear transfer is not a training-free adaptation: it provides an initialization strategy for better downstream performance, running with one auxiliary training objective along the fine-tuning.
Third, the interpretation in~\Cref{supp:transfer_explanation} is based on and consistent with the empirical observation, rather than a formal derivation.
A complete formal theory of Softmax-to-linear Transfer or Attention Transfer remains an important direction for future work.

\end{document}